\documentclass{article}
\usepackage{arxiv}

\usepackage[strings]{underscore}
\usepackage[utf8]{inputenc} 
\usepackage[T1]{fontenc}    
\usepackage{hyperref}       
\usepackage{url}            
\usepackage{booktabs}       
\usepackage{amsfonts}       
\usepackage{nicefrac}       
\usepackage{microtype}      
\usepackage{graphicx}
\usepackage[numbers,square]{natbib}
\usepackage{appendix}
\usepackage{amssymb}
\usepackage{amsmath,bm}
\usepackage{caption}
\usepackage[capitalise]{cleveref}
\usepackage{csquotes}
\usepackage{subcaption}
\usepackage{adjustbox}
\usepackage{underscore}
\usepackage{multirow}
\usepackage[dvipsnames]{xcolor}
\usepackage{graphicx}
\usepackage{siunitx}
\usepackage{xcolor}
\usepackage[font=small]{caption}
\usepackage{array}
\usepackage{enumitem}
\usepackage{titlesec}
\usepackage{fancyhdr}
\usepackage{natbib}
\usepackage{parskip}
\usepackage{mdframed}
\usepackage{listings}
\usepackage{algorithm}
\usepackage{algpseudocode}
\usepackage{tikz}
\usepackage{booktabs}
\usepackage{xcolor}
\usepackage{pifont}   
\usepackage{tikz}
\usepackage{rotating} 
\usetikzlibrary{positioning, shapes.geometric, arrows.meta}
\usetikzlibrary{shapes.geometric, arrows.meta, positioning, fit, backgrounds}

\definecolor{uqsaBlue}{RGB}{30, 80, 160}
\definecolor{uqsaGreen}{RGB}{20, 120, 60}
\definecolor{uqsaGray}{RGB}{80, 80, 90}
\definecolor{uqsaLight}{RGB}{240, 245, 255}
\definecolor{codebg}{RGB}{245, 247, 250}

\definecolor{cGreenFill}{RGB}{225,245,238}
\definecolor{cGreenBord}{RGB}{15,110,86}
\definecolor{cGreenText}{RGB}{8,80,65}
\definecolor{cPurpFill} {RGB}{238,237,254}
\definecolor{cPurpBord} {RGB}{83,74,183}
\definecolor{cPurpText} {RGB}{60,52,137}
\definecolor{cOranFill} {RGB}{250,236,231}
\definecolor{cOranBord} {RGB}{153,60,29}
\definecolor{cOranText} {RGB}{113,43,19}
\definecolor{cBlueFill} {RGB}{230,241,251}
\definecolor{cBlueBord} {RGB}{24,95,165}
\definecolor{cBlueText} {RGB}{12,68,124}
\definecolor{cGrayFill} {RGB}{241,239,232}
\definecolor{cGrayBord} {RGB}{95,94,90}
\definecolor{cGrayText} {RGB}{68,68,65}
\definecolor{cGoldFill} {RGB}{250,238,218}
\definecolor{cGoldBord} {RGB}{133,79,11}
\definecolor{cGoldText} {RGB}{99,56,6}
\definecolor{cBodyText} {RGB}{61,61,58}
\definecolor{cMuted}    {RGB}{115,114,108}

\definecolor{jsoncomment}{RGB}{130,130,130}
\definecolor{jsonbool}{RGB}{170,55,55}

\lstdefinelanguage{json}{
	basicstyle   = \ttfamily\footnotesize,
	columns      = fullflexible,
	keepspaces   = true,
	showstringspaces = false,
	breaklines   = true,
	morecomment  = [l]{//},
	commentstyle = \color{jsoncomment}\itshape,
	morekeywords = {true, false, null},
	keywordstyle = \color{jsonbool}\bfseries,
}

\tikzset{
	cpbox/.style={
		rounded corners=5pt,
		minimum width=2cm,
		minimum height=1.6cm,
		align=center,
		font=\small,
		line width=0.4pt,
	},
	cpbadge/.style={
		circle, inner sep=1pt,
		minimum size=0.55cm,
		font=\bfseries\scriptsize,
		text=white,
	},
	infobox/.style={
		rounded corners=5pt,
		align=center,
		font=\scriptsize,
		line width=0.4pt,
		inner sep=6pt,
	},
	seclabel/.style={font=\small\bfseries, text=cBodyText},
	muted/.style={font=\scriptsize, text=cMuted},
	blocking/.style={font=\scriptsize\itshape, text=cBodyText},
}

\newmdenv[
backgroundcolor=uqsaLight,
linecolor=uqsaBlue,
linewidth=1.5pt,
leftmargin=0pt,
rightmargin=0pt,
innerleftmargin=10pt,
innerrightmargin=10pt,
innertopmargin=8pt,
innerbottommargin=8pt
]{infobox}

\newtheorem{definition}{Definition}

\definecolor{henaBlue}{RGB}{30,80,160}
\definecolor{henaGreen}{RGB}{20,140,80}
\definecolor{henaOrange}{RGB}{200,100,20}
\definecolor{lightgray}{RGB}{245,245,245}

\title{Learning to Choose: An Empowerment-Guided Multi-Agent System with semantic communication for Adaptive Method Selection}

\author{ \href{https://orcid.org/0009-0002-4531-7960}{\includegraphics[scale=0.06]{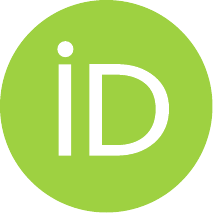}\hspace{1mm}Geremy Loachamin-Suntaxi\thanks{Authors also affiliated with the School of Chemical Engineering, National Technical University of Athens, Zographos Campus, 15780, Attiki, Greece}}\\
	Faculty of Science, Technology and Medicine\\
	University of Luxembourg\\
	Esch-sur-Alzette, L-4364, Luxembourg\\
	\texttt{geremy.loachamin@uni.lu}
	\And
    Robert Lazar\\
	Independent Researcher\\
	Arlington, VA, USA\\
	\texttt{tyrus400@gmail.com}
	\And
	\href{https://orcid.org/0000-0003-2272-2584}
	{\includegraphics[scale=0.06]{orcid.pdf}\hspace{1mm}Dimitrios G. Giovanis} \\
	Department of Civil \& Systems Engineering \\
	Department of Applied Mathematics and Statistics\\
	Johns Hopkins University\\	Baltimore, MD 21218, USA\\
	\texttt{dgiovan1@jhu.edu}\\
	\And
	\href{https://orcid.org/0000-0003-2220-3522}{\includegraphics[scale=0.06]{orcid.pdf}\hspace{1mm}Ioannis G. Kevrekidis} \\
	Department of Chemical and Biomolecular Engineering\\
	Department of Applied Mathematics and Statistics\\ Johns Hopkins University, USA\\
	Baltimore, MD 21218, USA\\
	\texttt{yannisk@jhu.edu}\thanks{Co-corresponding author: \texttt{yannisk@jhu.edu}}
    \And
    \href{https://orcid.org/0000-0002-5229-4157}{\includegraphics[scale=0.06]{orcid.pdf}\hspace{1mm}Eleni D. Koronaki}\\
	Luxembourg Institute of Science and Technology\\
	Esch-sur-Alzette, L-4362, Luxembourg\\
	\texttt{eleni.koronaki@list.lu}\thanks{Corresponding author: \texttt{eleni.koronaki@list.lu}}
}
\begin{document}
	
	\setcounter{page}{1}
	\maketitle
    \pagestyle{plain}
	\begin{abstract}
		Automating scientific computing workflows requires more than generating executable code: autonomous systems must also select appropriate computational strategies, implement them faithfully, and ensure that the resulting outcomes remain causally attributable to the decisions that produced them.
        In multi-agent pipelines, this process is particularly fragile, as small inconsistencies between agent 
        intents/actions
        can lead to semantic drift, where the eventual executed procedure no longer reflects the initially selected strategy, corrupting downstream evaluation and adaptation. 
        In this work, 
        motivated by the ATHENA framework \cite{toscano2025athenaagenticteamhierarchical,toscano2026graftathenaselfimprovingagenticteams}
        and the concept of empowerment in \cite{yiu2025empowerment}
        we introduce a multi-agent framework that combines contextual bandits with structured inter-agent communication and
        --most importantly--
        {\em semantic checkpoints} to preserve action–outcome fidelity throughout the pipeline. The system integrates specialized LLM agents, grounded code generation, and self-healing execution loops within an adaptive decision-making architecture. 
        Interpreting the framework through the lens of {\em empowerment}, we show that reliable autonomous learning requires both identifying high-quality actions {\em and} preserving the integrity of their propagation across agents. 
        Using sensitivity analysis and uncertainty quantification workflows as our representative illustrative case studies, we demonstrate that unchecked semantic drift degrades policy learning, while the proposed framework improves convergence, robustness and adaptation to novel problem contexts. 
        These results suggest a general design principle for scientific multi-agent systems: adaptive decision-making must be coupled with explicit mechanisms that guarantee semantic consistency and reliable information flow.
	\end{abstract}
	
	
	\section{Introduction}\label{sec:Introduction}	
	
	The automation of scientific computing workflows presents a fundamental challenge that goes beyond code generation \cite{lyu2026understandingbridgingplannercodergap}. \textcolor{black}{As recently illustrated in practice \cite{langer2026communication}, a "Claude Code Hacker" identified that failures can already occur at the very first step, when an agent silently disregards the context the author intended and proceeds on a self-fabricated problem statement, invalidating downstream evaluation regardless of how well subsequent actions are chosen.} An autonomous system must not only produce executable code, but select, from a growing library of methods, the one most likely to yield reliable results for the problem at hand, and ensure that this selection survives intact through every stage of a multi-agent pipeline. 
    This \emph{selection-to-execution} problem has an inherently sequential character: information gathered by running one method informs the next choice and is precisely the setting that contextual bandit theory \cite{bubeck2012regret}
    was designed to address. 
    
    However, method selection alone is insufficient. Even when a policy identifies the correct action for a given problem, that choice must propagate through the pipeline without distortion before the resulting reward can provide a valid estimate of the action--outcome relationship.
    In a multi-agent pipeline, success of this propagation is not guaranteed: small semantic distortions accumulate across agent transitions, so that the code executed may no longer reflect the method that was initially selected, and the reward observed measures {\em code} quality rather than {\em method} quality. This \emph{semantic drift}, the progressive decoupling of what the policy chose from what the executor ran, arises precisely because selection and implementation are assigned to distinct agents, a separation that is necessary for the bandit to receive a clean learning signal but that introduces new failure modes at every inter-agent boundary.
	
	A principled channel through which to understand both the selection and propagation problem is the notion of \emph{empowerment} \citep{klyubin2005empowerment, salge2014empowerment}. As argued by \citet{yiu2025empowerment} in the context of human learning, an agent that maximizes mutual information between its actions and the resulting state changes, can be thought of as simultaneously discovering the controllable, causal structure of its environment. 
    %
    In a scientific computing pipeline, high empowerment means that method choices reliably determine computational outcomes: the system is controllable. Low empowerment means that actions fail to faithfully propagate: semantic drift between agents, implementation errors, or failures in inter-agent communication degrade the action--outcome link, so that even correctly selected actions may no longer produce reliably attributable outcomes.
    This decomposition reveals that high system empowerment requires two distinct mechanisms simultaneously: (a) a contextual bandit policy that identifies which actions causally produce high-quality results for a given problem context, and (b) explicit guarantees of good communication, that ensure those actions are faithfully preserved as they propagate through code generation and execution.
	
	The present work explores this general principle in a multi-agent system for the illustrative task of automated sensitivity analysis (SA) and uncertainty quantification (UQ). SA and UQ are key tasks in the simulation of physical, engineering, and scientific models \cite{uqpy2022, UQLab, Saltelli2010}. Despite a rich theoretical literature and mature software libraries, their practical application remains demanding: the analyst must select an appropriate method, tune hyperparameters, implement the workflow in code, and interpret the results, steps that require domain expertise rarely co-located with the modeling effort. Large language models have recently demonstrated the ability to generate, debug, and explain scientific code \cite{chen2021codex}, and multi-agent architectures extend this capability by decomposing complex tasks into specialized sub-agents \cite{anthropic_agents2024, swebench2023, schmidgall2024, Swanson2024.11.11.623004, cheng2023}.
    Our inspiration comes from ATHENA \cite{toscano2025athenaagenticteamhierarchical}, an agentic platform that has demonstrated how a contextual-bandit-driven framework can manage the end-to-end research lifecycle by treating structural design choices as bandit actions and using observed performance as feedback. \textcolor{black}{To maintain scientific foundation throughout its pipeline, ATHENA employs expert-curated {\em blueprints}: compact representations of domain knowledge covering approximation theory, physics-informed constraints, and numerical methods that guide agent reasoning within established scientific principles. The present work takes a complementary approach: rather than reducing semantic drift primarily through expert knowledge injection, we detect and correct it through {\em semantic checkpoints} placed at critical boundaries between agents. 
    This distinction arises because ATHENA operates in settings where many scientific constraints and workflow structures can be specified \emph{a priori}, whereas our pipeline must remain largely method-agnostic: the generated implementations may correspond to substantially different analysis methods, each with distinct assumptions, output schemas, computational characteristics, and associated reward behavior within the bandit loop.
     Checkpoints are therefore not an alternative to expert knowledge, but rather a structural complement that preserves semantic coherence across agent transitions regardless of the method being executed, catching mismatches that neither domain knowledge nor prompt engineering can fully anticipate.}

     Our inspiration comes from ATHENA \cite{toscano2025athenaagenticteamhierarchical}, an agentic platform that has demonstrated how a contextual-bandit-driven framework can manage the end-to-end research lifecycle by treating structural design choices as bandit actions and using observed performance as feedback. \textcolor{black}{More recent extensions of this direction, particularly GRAFT-ATHENA \cite{toscano2026graftathenaselfimprovingagenticteams}, further reinterpret scientific automation as a persistent learning system operating over structured problem and method spaces rather than as a sequence of isolated optimization episodes. In GRAFT-ATHENA, previously solved problems and successful computational strategies are embedded into a geometric memory through graph-based factorizations and methodological fingerprints, enabling cross-problem transfer, warm-started exploration, and continual accumulation of scientific experience. This introduces a complementary perspective to the present work: whereas GRAFT-ATHENA focuses on the geometric organization and transfer of methodological knowledge across sessions, we focus on preserving the semantic integrity of action propagation within a single multi-agent execution trajectory. In particular, we investigate how semantic drift between agents can corrupt the causal relationship between selected methods and observed computational outcomes, thereby degrading the learning signal used by the contextual bandit policy.}

\textcolor{black}{To maintain scientific foundation throughout its pipeline, ATHENA employs expert-curated blueprints: compact representations of domain knowledge covering approximation theory, physics-informed constraints, and numerical methods, that guide agent reasoning within established scientific principles. In the more recent GRAFT-ATHENA extension, these structured priors evolve into persistent methodological representations embedded within graph-based problem and action spaces, enabling cross-problem transfer, reusable computational motifs, and continual accumulation of scientific experience. The present work takes a complementary approach: rather than addressing semantic drift primarily through structured methodological priors, knowledge injection, and persistent scientific memory, we detect and correct it through semantic checkpoints placed at critical boundaries between agents. This distinction reflects the nature of the two settings: ATHENA and GRAFT-ATHENA rely on the ability to encode substantial portions of scientific structure in advance, whereas our pipeline must remain agnostic to the specific method being executed, whose output schema and numerical invariants vary across estimators. Checkpoints are therefore not an alternative to expert knowledge but a structural complement that verifies inter-agent coherence regardless of the method, catching mismatches that neither domain knowledge, persistent memory, nor prompt engineering can fully anticipate.}

	Inspired by ATHENA, our approach also frames method selection as a contextual bandit problem: at each iteration, a context vector $\mathbf{x}_n$ is constructed that encodes the characteristics of the problem; an action $a_n$ is chosen by a policy $\pi$; the corresponding code is generated and executed; and a scalar reward $R_n$ is computed. The policy is updated online using the observed rewards, so that future selections reflect previously observed action--outcome performances.
    Method selection via multi-armed bandits has been studied in the AutoML literature \cite{thornton2013, feurer2015}. Contextual bandits extend this by conditioning the policy on observable features \cite{langford2007, toscano2025athenaagenticteamhierarchical}. Thompson Sampling \cite{thompson1933} provides a principled exploration-exploitation trade-off with strong theoretical guarantees \cite{agrawal2013}. In this work, we introduce a multi-agent framework for SA/UQ method selection that combines contextual bandits with structured mechanisms to mitigate semantic drift, incorporates cross-session warm-starting to accelerate adaptation on familiar problems, and interprets the resulting system through the lens of empowerment from reinforcement learning.
    
    We introduce three architectural components that realize this decomposition of empowerment: The first is a hierarchy of \emph{structured schemes}: a problem scheme, a method scheme, and a diagnostic scheme that replace free-text inter-agent communication with machine-readable structured schemes at every critical boundary, eliminating the ambiguity that would otherwise allow a correct policy decision to be silently overridden or erroneously implemented downstream. The second is {\em a set of cosine-similarity checkpoints} placed at specific agent transitions, each targeting a distinct failure mode in the causal chain from selection to reward. Together, schemes and checkpoints implement the communicative component of system empowerment, ensuring that the controllable action-outcome pairs discovered by the bandit are preserved as the pipeline executes. The third contribution is a cross-session context checkpoint that compares the current problem against an archive of completed sessions: on a familiar problem, it warm-starts the policy, accelerating convergence; on a structurally novel problem, it activates an \textit{anomaly-directed exploration strategy}, maximizing variability before committing to an estimator.
	
	The work is structured as follows. Section \ref{sec:problem} develops the theoretical foundations, opening with the empowerment concept and its role as a coordination principle within the multi-agent framework, before formalizing the contextual bandit setting that supports method selection. Section \ref{sec:agents} presents the multi-agent architecture and the semantic checkpoint mechanism. Section \ref{sec:experiments} reports the experimental results of the planned benchmark test cases. Section \ref{sec:conclusion} concludes with a discussion of implications and future work. The details of the implementation are collected in the Appendix \ref{sec:details}.
	
	
	\section{Methods}\label{sec:problem}

	\subsection{The Empowerment Concept}
	
	Autonomous agents that learn by interacting with an environment face a fundamental tension between exploiting known actions and exploring new ones. Empowerment \citep{klyubin2005empowerment} resolves this by providing an intrinsic reward that does not measure the value of the outcomes, but rather how strongly the agent's actions determine them.
	
	Concretely, an agent seeking empowerment simultaneously pursues two properties. \emph{Controllability}: its actions reliably produce distinct, predictable outcomes. \emph{Variability}: it explores a wide range of actions rather than converging on a single one. \citet{yiu2025empowerment} showed empirically that this combination is how humans build useful causal models of their environment, and that the relationship is bidirectional: an agent with a better causal model exercises greater control, which increases empowerment, which in turn drives further causal learning.
	
	Formally, for an agent in state $s$ that can take an action $a \in \mathcal{M}$ leading to a successor state $s'$, \textit{empowerment} (Empw) is defined as the channel capacity through which the agent guides its environment:
	\begin{equation}
		\text{Empow}(s) \ =\  \max_{p(a)} \  I\ \left(\mathbf{A};\  s' \mid s\right)
		\label{eq:empowerment}
	\end{equation}
	
	where $I(\mathbf{A}; s' \mid s)$ is the mutual information between the action random variable $\mathbf{A}$ and the successor state $s'$, maximized over all possible action distributions $p(a)$.
	
	\medskip
	
	In a multi-agent framework, the empowerment principle has a direct operational meaning. At each iteration, the system must choose an action from the available action space. Whether that choice is empowering depends on how reliably the selected action causally produces an outcome, one that is directly attributable to the action taken, rather than to external factors outside the agent's control. An action that consistently delivers convergent, interpretable results exercises strong causal influence over the outcome and is therefore empowering. An action that frequently fails, produces degenerate results, or is simply inappropriate for the context does not.
	
	Reliable action selection is, however, necessary but not sufficient. The pipeline operates as a chain of agents in which the output of one becomes the input to the next. For the initial decision to translate into a high-quality result, it must survive intact through every agent to agent transition. If any agent misinterprets, distorts, or ignores the information passed to it, the causal link between the action chosen and the outcome produced is broken, regardless of how well the action was initially selected. This structural fragility motivates a system-level definition of empowerment that captures not just the quality of the action chosen, but the integrity of its propagation through the pipeline. 
	
	In the context of multi-agent scientific computing systems (e.g., for automated sensitivity analysis and uncertainty quantification), this pipeline translates a method selection, an element of the combinatorial action space $\mathcal{M}$, into a computational outcome and diagnostics. Each step is conditioned on a fixed \textit{context vector} $\mathbf{x}$ that summarizes the structural characteristics of the problem (e.g., input and output dimensionality, sampling budget, distributional family, and feasibility constraints). This vector describes the problem, not the state of the system at a particular iteration; it is computed once from the problem description and remains constant throughout execution. The system empowerment associated with a problem instance described by $\mathbf{x}$ is then defined as follows.
	
	\begin{definition}[System Empowerment]\label{def:system_empowerment}
		Let $\mathcal{S}$ denote a multi-agent pipeline: an ordered composition of implementation and execution operators that translates a method selection into a computational outcome. Let $f$ be a computational model, $\mathcal{M}$ a finite set of available methods, and let $\mathbf{A}$ and $\mathbf{O}$ be random variables representing, respectively, the method-selection action taking values in $\mathcal{M}$, and the computational outcome produced by executing $\mathbf{A}$ on $f$ through $\mathcal{S}$. The full state of $\mathcal{S}$ at any decision point is $s = (\mathbf{x}, \mathcal{H})$, where $\mathbf{x}$ is the context vector and $\mathcal{H}$ is the history of previous actions and outcomes accumulated up to that point. The successor state $s'$ is identified with $\mathbf{O}$: conditioned on $s$, $\mathbf{O}$ is the only component that varies with the action choice. The \emph{system empowerment} of $\mathcal{S}$ at state $s$ is therefore:
		$$\emph{Empow}_{\mathcal{S}}(s) = I\!\left(\mathbf{A};\, \mathbf{O} \mid s\right).$$
	\end{definition}

	This quantity measures how strongly method choices determine computational outcomes for a problem described by $\mathbf{x}$. High empowerment means that method selections reliably determine computational outcomes: the pipeline is controllable. Low empowerment means that actions fail to propagate: semantic drift between agents, implementation errors, or failures in inter-agent communication degrade the action-outcome link, making the workflow uncontrollable regardless of how well the method was selected.
	
	Definition \ref{def:system_empowerment} reveals that high system empowerment requires two distinct mechanisms. The first is a policy that learns which method selections causally produce high-quality results for a given problem context, the \emph{statistical} component of $\text{Empow}$. The second is a mechanism that ensures the selected action is faithfully preserved as it propagates through the agent pipeline, the \emph{communicative} component of $\text{Empow}$. Together, they implement a separation of concerns between choosing the right action and ensuring it propagates. Both are necessary conditions for high system empowerment.
	
	This suggests a general organizing principle for multi-agent scientific computing systems: \emph{pair an exploration policy with communicative guarantees}. The exploration policy expands the set of controllable action-outcome pairs the system can reliably exploit; the communicative guarantees ensure that this controllability is not lost in translation between agents. Every architectural decision in the framework presented in this paper follows from this decomposition.
	
	Definition \ref{def:system_empowerment} has a direct architectural interpretation. The mutual information $I(\textbf{A}; \textbf{O} \mid s)$ decomposes naturally into two components that map onto two distinct groups of the system. The first is statistical: the system maximizes $I(\textbf{A}; \textbf{O} \mid s)$ by learning, through the contextual bandit loop, which method selections reliably produce high-quality outcomes for a given problem context. The second is communicative: semantic checkpoints preserve this mutual information by ensuring that the selected action $\textbf{A}$ survives intact through code generation and execution. If \textit{semantic drift} decouples $s$ from $\textbf{A}$, the observation $\textbf{O}$ becomes statistically independent of the action that nominally produced it, reducing $I(\textbf{A}; \textbf{O} \mid \mathbf{x})$ to zero regardless of how well the policy was selected/updated.

	\subsection{Contextual Bandit Setting}
	
	The statistical component of system empowerment, i.e. learning which method selections causally produce high-quality results, is formalized as a contextual bandit problem, following the formulation introduced in \cite{toscano2025athenaagenticteamhierarchical}. The system manages a finite set $\mathcal{M} = \{m_1, \ldots, m_k\}$ of available methods, which constitute the arms of the bandit. At each iteration $n \in \{1, \ldots, N_{\max}\}$, the system observes the context $\mathbf{x} \in [0,1]^d$ encoding the problem characteristics, selects a method $A_n \in \mathcal{M}$ according to a policy $\pi$, and generates an executable program $S_n$ that instantiates $A_n$ for the selected method on $f$. The execution of $S_n$ produces an observation $O_n$, comprising in our illustrative example sensitivity indices or uncertainty estimates, convergence diagnostics, and plots, from which a scalar reward $R_n$ is computed measuring the quality of the result. 
	
	The triple $(A_n, S_n, O_n, R_n)$ is appended to the history $\mathcal{H}_n = \{(A_i, S_i, O_i, R_i)\}_{i=1}^{n}$, which accumulates evidence about which methods work well for the problem described by $\mathbf{x}$.
	
	Formally, the policy $\pi = \bigl(\pi_n(A \mid \mathbf{x}, \mathcal{H}_{n-1})\bigr)_{n \in \mathbb{N}}$ defines a probability distribution over arms, conditioned on both the fixed problem context $\mathbf{x}$ and the full history $\mathcal{H}_{n-1}$ of all previous method selections, executable implementations, observations, and rewards. The context encodes what the problem is and the history encodes what has been learned about it across all previous iterations. Thus, for a horizon $N \geq 1$, the objective is to minimize cumulative regret:
	$$\mathrm{Regret}_\pi(N) = \mathbb{E}_\pi\ \left[\sum_{n=1}^{N} R_n^* - R_n \right],$$
	
	where $R_n^*$ is the expected reward of the optimal action given the context. A policy is said to \emph{learn} if $\mathrm{Regret}_\pi(N) = o(N)$, meaning cumulative regret grows sublinearly and the gap to the expected reward of the optimal action decreases over time.
	
	In the finite-horizon setting of a single session, where $N_{\max}$ is typically small, a stronger condition is desirable. We therefore additionally require that the policy satisfy a \emph{submartingale} convergence condition:
	$$\mathbb{E}\ \left[R_{n+1} \mid \mathcal{H}_n\right] \ \geq\  R_n, \qquad \forall\ n \in \mathbb{N}.$$
	This condition states that, given the history accumulated, the expected reward at the next iteration is at least as large as the current one. It is the formal statement that the system is improving monotonically in expectation, not merely on average over the full run. Unlike the asymptotic regret bound, this condition is verifiable at runtime after each iteration and provides a practical convergence criterion independent of $R_n^*$.

	\section{Multi-Agent Architecture}\label{sec:agents}
	
	The multi-agent system is organized into four functional groups. The \textit{Conceptualization Team} runs once per problem and extracts a structured representation of the user's request; it consists of the \texttt{Coordinator}, \texttt{Gatekeeper}, and \texttt{Model Translator}. The \textit{Strategy Team}, consisting of the \texttt{Strategist}, \texttt{Critic}, and \texttt{Advisor}, functions as the policy operator $\pi$, selecting and evaluating methods based on context and history. The \textit{Implementation Team}, includes the \texttt{Study Agent}, \texttt{Refactor Agent}, and \texttt{Inspector}, functions as the implementation operator $\mathcal{I}$, translating abstract method selections into executable code. The \textit{Execution Team}, comprising the \texttt{Execution Runner} and \texttt{Debugger}, functions as the execution operator $\mathcal{E}$, running the generated code and recovering from runtime errors to produce results, plots, and diagnostic reports. The \texttt{Advisor} functions as the evaluation operator $\mathcal{A}$, analyzing the observation $O_n$ and assigning the scalar reward $R_n$ together with a diagnosis and prescribed correction for the next iteration; it is the only multimodal agent in the pipeline, receiving both numerical results and plots. 
	
	The Register closes the loop by appending $(A_n, S_n, O_n, R_n)$ to the history $\mathcal{H}_n$ and verifying the submartingale condition $R_n \geq R_{n-1}$. If the condition is violated, it is recorded in $\mathcal{H}_n$ and passed to the \texttt{Strategist} at the next iteration as a structured signal, informing the proposal of $A_{n+1}$ with the diagnosis that the previous method underperformed relative to its predecessor. This makes the submartingale condition operationally verifiable at runtime rather than merely asymptotically: after each iteration the system has explicit evidence of whether it is improving, and that evidence feeds directly into the next selection decision through the \texttt{Advisor}'s, which the \texttt{Strategist} reads alongside the full history $\mathcal{H}_{n-1}$.
	
	Thus, at each iteration $n$ of the contextual bandit loop, the system executes the following sequential update:
	\begin{align*}
		A_{n+1} &\sim \pi\ \left(A_{n+1} \mid \mathbf{x},\ \mathcal{H}_{n}\right) &\qquad \text{(policy)}\\
		S_{n+1} &\sim \mathcal{I}\ \left(A_{n+1},\ S_{n},\ \mathcal{D}\right) &\qquad \text{(implementation)}\\
		O_{n+1} &\sim \mathcal{E}\ \left(O_{n+1} \mid S_{n+1}\right) &\qquad \text{(execution)}\\
		R_{n+1} &\sim \mathcal{A}\ \left(R_{n+1} \mid O_{n+1}\right) &\qquad \text{(evaluation)}\\
		\mathcal{H}_{n+1} &\leftarrow\ \mathcal{H}_{n} \cup \{(A_{n+1},\ S_{n+1},\ O_{n+1},\ R_{n+1})\} &\qquad \text{(register)}
	\end{align*}
	
	The \texttt{Inspector} agent verifies that $S_{n+1}$ faithfully performs the 
	requested action $A_{n+1}$ by enforcing a structural checklist (e.g., correct UQpy class implementation, parameter matching, and output 
	completeness), mitigating the influence of $S_n$ on $S_{n+1}$ and 
	approximating the conditional independence $S_{n+1} \perp S_n \mid 
	A_{n+1}$ required by the bandit formulation. This cycle repeats until 
	$R_n \geq R_{\mathrm{threshold}}$ or the maximum number of iterations is reached, $n = N_{\max}$.

    \subsection{Problem Formulation in Sensitivity Analysis and UQ}
	
	Let $f : \mathcal{X} \subseteq \mathbb{R}^{d_{\mathrm{in}}} \to \mathbb{R}^{d_{\mathrm{out}}}$ be a computational model whose inputs $\mathbf{X} = (X_1, \ldots, X_{d_{\mathrm{in}}})$ are uncertain random variables with joint distribution $F_\mathbf{X}$. Given a computational budget of $N$ model evaluations and a precision target $\varepsilon$, the goal is to \emph{characterize} the uncertainty structure of $f$ under $F_\mathbf{X}$. This includes two complementary tasks:
	
	\begin{itemize}[leftmargin=2em, itemsep=4pt]
		\item \textbf{Sensitivity Analysis:} quantify how much each input $X_i$ individually or in interaction with other inputs, contributes to the variability of the output. This includes first-order effects, which only consider the contribution of each $X_i$ alone, and higher-order interaction effects, which capture joint contributions of subsets of inputs. The result is a set of sensitivity measures (e.g. Sobol indices, Morris elementary effects, or Chatterjee coefficients). When analytical reference values are available (as in standard benchmark functions such as the Ishigami function), the quality of the estimates can be assessed directly by comparison. In general practice, convergence is instead monitored across successive estimates as the sample size grows.
		
		\item \textbf{Uncertainty Quantification:} estimate distributional properties of the output $Y = f(\mathbf{X})$, including its mean, variance, confidence intervals, and tail probabilities. As with SA, the accuracy of UQ estimates is verifiable only when closed-form reference values are available; otherwise convergence serves as the practical quality criterion.
	\end{itemize}
	
	Both tasks share the same inputs: a model $f$, a distribution $F_\mathbf{X}$, a budget N, and a precision target $\varepsilon$. They differ in the quantity of interest they seek to estimate, and consequently in the class of methods appropriate for each. We denote the task type by
	$\tau \in \{\mathrm{SA}, \mathrm{UQ}\}$.
	
	Both tasks share a common challenge: the analyst must choose, from a growing library of methods, the one most likely to produce reliable, interpretable results for the specific problem at hand. A weak choice, applying a variance-based method to a model with too few samples, or a screening method to a low-dimensional problem, wastes the computational budget and yields estimates of little value. A strong choice, by contrast, produces results that are directly controlled by the method applied: the action determines the outcome. This relationship between actions and outcomes, the degree to which method choices causally govern the quality of the uncertainty characterization, is the central learning problem that the multi-agent framework is designed to solve, and it is precisely the notion we formalize in the next section.

	\subsection{System Design}\label{sec:design}
	
	The multi-agent framework is organized around three design decisions that jointly determine whether the bandit loop can learn anything useful.
	
	The first decision is the definition of the action space $\mathcal{M}$. Method selection is only learnable if the actions are finite and distinguishable. Rather than a flat set of methods, the multi-agent framework defines separate combinatorial action spaces for SA and UQ tasks. The SA space is structured along four dimensions: estimator family, surrogate model, budget allocation strategy, and output treatment scheme, so that $\mathcal{M}_{\text{SA}} = \mathcal{D}_1 \times \mathcal{D}_2 \times \mathcal{D}_3 \times \mathcal{D}_4$. The UQ space extends this to six dimensions, adding uncertainty propagation method and reliability analysis strategy, giving $\mathcal{M}_{\text{UQ}} = \mathcal{D}_1 \times \cdots \times \mathcal{D}_6$. In both cases, each element $A = (d_1, \ldots, d_k)$ specifies a complete analytical pipeline. Each space is further restricted at every iteration to a feasible subset $\mathcal{M}_n \subseteq \mathcal{M}$ by filtering action vectors that violate hard constraints derived from $\mathbf{x}$, for example, $\mathcal{D}_2 = \texttt{Morris}$ is blocked when $d_{\mathrm{in}} < 8$ (see Appendix \ref{sec:details}). Cross-dimensional constraints reduce both spaces to a context-dependent feasible set, preserving finiteness and distinguishability.
	
	The second decisoin is the reward signal $R_n$. For the bandit to converge, the reward must be attainable by some policy. The multi-agent framework  defines $R_n$ as a composite of four components capped at physically meaningful thresholds:
	$$R_n = R_{\mathrm{integrity}} + R_{\mathrm{accuracy}} + R_{\mathrm{details}} + R_{\mathrm{optimality}},$$
	
	where $R_{\mathrm{integrity}}$ measures executability and correct LLM usage; $R_{\mathrm{accuracy}}$ evaluates convergence of indices and consistency constraints; $R_{\mathrm{details}}$ penalizes local failures; and $R_{\mathrm{optimality}}$ rewards computational efficiency. This design ensures that maximal reward $R_n$ is reachable, which is the necessary condition for vanishing regret.
	
	
	\begin{table}[ht]
		\centering
		\caption{Reward components, maximum values, and evaluation criteria.}
		\label{tab:reward}
		\begin{tabular}{llp{7cm}}
			\toprule
			\textbf{Component} & \textbf{Max} & \textbf{Criteria} \\
			\midrule
			$R_{\mathrm{integrity}}$ & 35 & Code executed without critical errors; UQpy API invoked correctly; all expected outputs present. \\
			$R_{\mathrm{accuracy}}$ & 35 & $S_{\mathrm{prec}}$: exact MAE against analytical reference when available, convergence criterion ($|\hat{S}_i^n - \hat{S}_i^{n-1}| < 10^{-3}$); $S_{\mathrm{cons}}$: consistency constraints $\big(\sum S_i \in [0,1]; \ ST_i \geq S_i\big)$. \\
			$R_{\mathrm{details}}$ & 15 & Penalty $-5$ per NaN index; $-5$ for negative variance; $-3$ for rank inversion between iterations; $-2$ for unaddressed warnings. \\
			$R_{\mathrm{optimality}}$ & 15 & $N_{\mathrm{used}}$ vs.\ theoretical minimum; runtime vs.\ UQpy baseline; memory footprint. \\
			\bottomrule
		\end{tabular}
	\end{table}
	
	
	The third decision is the preservation of the causal chain $A_n \to S_n \to O_n \to R_n$. Even a well-designed action space and reward are useless if the signal from $A_n$ does not survive code generation and execution. If $S_n$ does not faithfully implement what was selected, then $R_n$ measures code quality rather than method quality, and the bandit learns nothing about methods. In the language of empowerment \citep{klyubin2005empowerment}, high system empowerment requires that action choices reliably determine outcomes. Each operator in the sequential update plays a distinct role in maintaining this chain:
	
	The \textbf{policy} $\pi(A_n \mid \mathbf{x}, \mathcal{H}_{n-1})$ is the selection action.

    It maps the fixed problem context $\mathbf{x}$ and the accumulated history $\mathcal{H}_{n-1}$ to a distribution over $\mathcal{M}_n$, implemented by the \textit{Strategy Team}. It implements \textit{controllability:} weighting selections by learned reward, and \textit{variability:} maintaining an exploration. The exploration-exploitation balance is managed through a decaying confidence schedule: early iterations operate in near-uniform exploration mode, diversifying method selections across $\mathcal{M}_n$ to discover controllable action-outcome pairs; later iterations transition smoothly toward exploitation, concentrating on configurations that have demonstrated high reward. This directly implements the variability requirement of system empowerment without an abrupt warm-up switch.
	

Moreover, to make the policy tractable in the combinatorial action space $\mathcal{M}$, each action vector $(d_1,\ldots,d_7)$ is encoded together with the problem context into a single feature vector, and a single linear reward model is learned over this combined representation. This avoids maintaining a separate reward estimator for every possible pipeline configuration, which would require observing each combination many times before reliable estimates could be formed. Instead, pipelines that share similar structural features also share model parameters. As a result, observing the reward from one pipeline updates not only that specific configuration, but also influences the estimated rewards of structurally related pipelines. In this way, information gathered from previously evaluated pipelines can inform the estimated reward of new pipeline configurations that share similar methodological components or structural characteristics.

	The \textbf{implementation operator} $\mathcal{I}(A_n, S_{n-1}, \mathcal{D})$ translates the abstract action $A_n$ into an executable program $S_n$, conditioned on the previous code state $S_{n-1}$ and the UQpy corpus $\mathcal{D}$. It is implemented by the Implementation Team through corpus-grounded retrieval and cell-by-cell refactoring. The \texttt{Inspector} agent strictly verifies that $S_n$ faithfully realizes $A_n$ regardless of $S_{n-1}$, which guarantees the Markov property $S_n \perp S_{n-1} \mid A_n$ required by the bandit formulation.
	
	The \textbf{execution operator} $\mathcal{E}(O_n \mid S_n)$ runs $S_n$ and produces the multimodal observation $O_n$ (e.g., sensitivity indices, convergence diagnostics, and plots). It is a deterministic function of $S_n$ implemented by the execution runner.

	The \textbf{evaluation operator} $\mathcal{A}(R_n \mid O_n, \mathbf{x}, \mathcal{H}_{n-1})$ maps the observation to a scalar reward $R_n$ and a diagnostic report. It is implemented by the \texttt{Advisor} agent, the only multimodal agent in the pipeline, and conditions on both the current observation and the problem context to assess convergence relative to the precision target $\varepsilon$.
	
	
	Inter-agent communication is structured through three \textbf{scheme} types that replace free-text fields as the primary carrier of information across the pipeline. A \textit{problem scheme} is constructed from the context vector before the bandit loop begins, encoding the pre-computed feasible method set and the dimensional classification of the problem; it conditions every subsequent agent decision and eliminates the need for agents to re-derive problem structure from raw text. A \textit{method scheme} is attached to each policy decision and carries the complete methodological specification of the selected action: the estimator family, the sensitivity indices it produces, its internal sampling scheme, and the minimum evaluation budget derived from closed-form formulas. A \textit{diagnostic scheme} is produced by the evaluating agent at the end of each iteration; its entries belong to a controlled vocabulary covering root cause, bottleneck dimension, prescribed estimator change, and prescribed budget multiplier, replacing the free-text diagnosis field with a directly machine-readable scheme. Together, these three structures make inter-agent signals clear and parseable, reducing the probability of semantic drift at every agent boundary.
	
	The multi-agent framework preserves the integrity of this chain through the UQpy corpus, which prevents hallucinated LLM calls from decoupling $S_n$ from $A_n$, and through the semantic checkpoints, which verify action-to-context coherence, code-to-action coherence, and diagnosis-to-observation coherence respectively, rejecting transitions where semantic drift would otherwise break up the connection between what the policy selected and what the reward measures.

\begin{sidewaysfigure}
    \centering
    \includegraphics[height=0.65\textwidth]{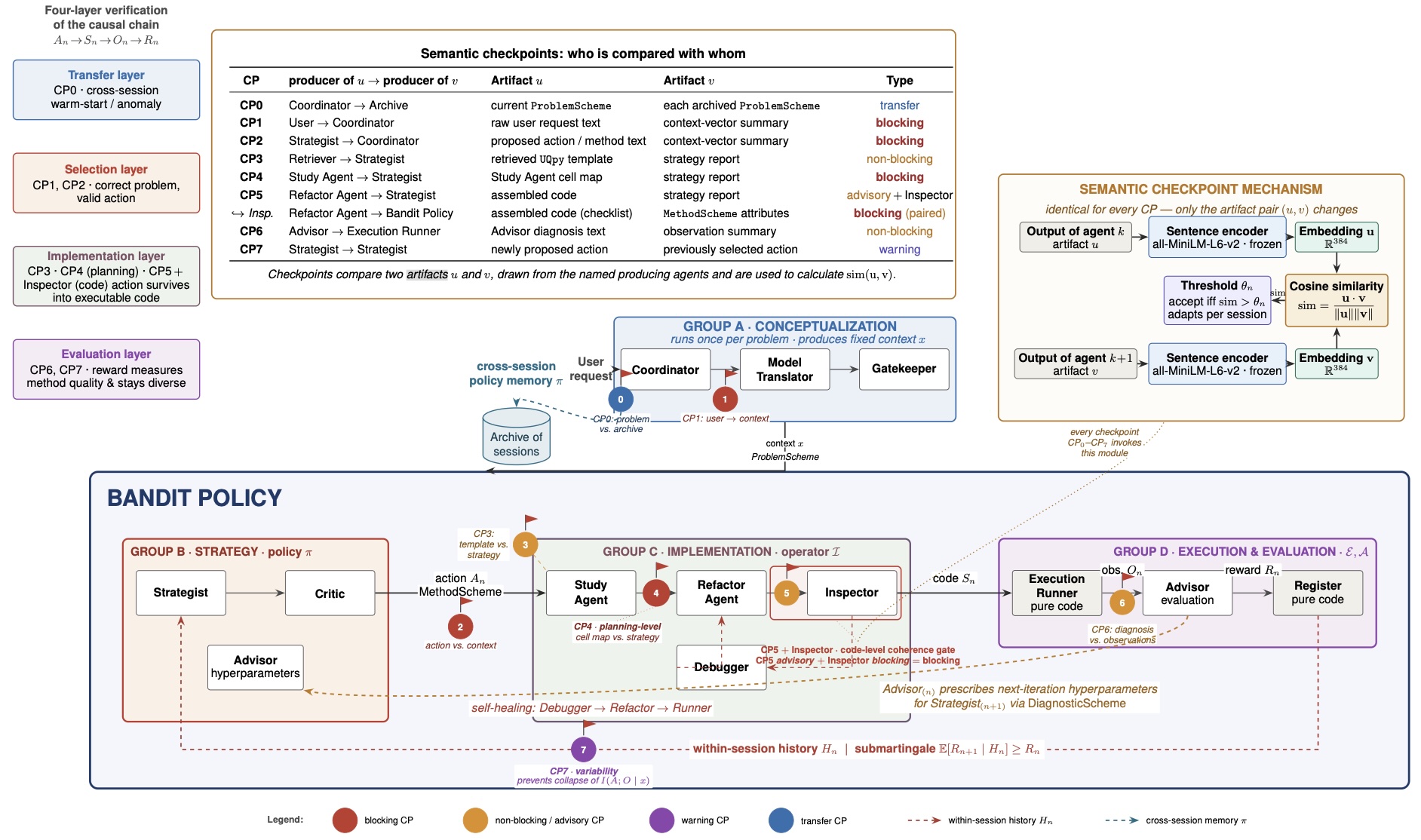}
    \caption{
Empowerment-stabilized multi-agent architecture for automated sensitivity analysis (SA) and uncertainty quantification (UQ). The figure illustrates a contextual bandit loop protected by seven semantic checkpoints (CP0--CP7) organized across four verification layers: Transfer, Selection, Implementation, and Evaluation. The Conceptualization stage transforms the user request into a structured problem representation and context vector, while CP0 compares the problem against a historical archive to distinguish familiar tasks from anomalous ones. The Strategy Team proposes candidate methods and verifies context compatibility before implementation. The Implementation Team translates the selected strategy into executable code through planning, template retrieval, code generation, inspection, and debugging stages, with CP4 and CP5 acting as a two-stage coherence gate to preserve semantic fidelity between the intended method and the assembled implementation. The Execution and Evaluation layer generates multimodal observations, including numerical results and plots, which are analyzed by the Advisor to produce grounded rewards and diagnostic feedback. CP6 ensures reward grounding in actual observations, while CP7 promotes action diversity within the contextual bandit policy. The Register updates the interaction history and enforces the submartingale condition $\mathbb{E}[R_n \mid H_{n-1}] \geq R_{n-1}$, encouraging monotonic improvement of the autonomous workflow over successive iterations.
}
    \label{fig:multi}
\end{sidewaysfigure}

	\subsection{System Architecture}\label{sec:system}
	
	The complete framework is organized as an empowerment-maximizing pipeline across two phases and four logical groups. At the highest level, the architecture implements the separation identified in \cref{def:system_empowerment}, the \textit{Strategy Team} maximizes the \textit{statistical} component of $\mathcal{E}(\mathbf{x})$ by learning which methods causally produce high-quality results; the checkpoints and scaffold preserve the \textit{communicative} component by ensuring those choices reliably propagate through code generation and execution. Table \ref{tab:agents} and Fig. \ref{fig:multi} summarize the full assignment.

	\subsubsection*{Group A: Conceptualization Phase} This phase runs once per problem and produces the fixed context $\mathbf{x}$ that conditions all subsequent decisions. The \texttt{Coordinator} parses the user's description and extracts the structured problem representation: input dimensionality $d_{\mathrm{in}}$, output dimensionality $d_{\mathrm{out}}$, sampling budget $N$, precision target $\varepsilon$, input distributions, and task type (SA or UQ). The output of the phase is the context vector $\mathbf{x} \in \mathbb{R}^8$, which encodes the structural characteristics of the problem instance as a fixed numerical representation held constant throughout the session. Concretely, $\mathbf{x}$ comprises: input dimensionality $d_{\mathrm{in}}$, output dimensionality $d_{\mathrm{out}}$, sampling budget $N$, precision target $\varepsilon$, task type $\tau \in \{\mathrm{SA}, \mathrm{UQ}\}$, distribution family (e.g., Gaussian, uniform, etc.), a multioutput indicator $\mathbf{1}_{d_{\mathrm{out}} > 1}$, and binary feasibility conditions encoding which methods in $\mathcal{M}$ satisfy the hard constraints derived from the problem description.
	
	The \texttt{Gatekeeper} validates this representation, routes the task to the appropriate combinatorial space $\mathcal{M}_{\text{SA}}$ or $\mathcal{M}_{\text{UQ}}$ according to the task type $\tau \in \{\text{SA}, \text{UQ}\}$, and enforces the feasibility constraints that define $\mathcal{M}_n$ before the policy is ever consulted. The \texttt{Model Translator} converts the user's mathematical expression into a compatible Python callable, handling the signature requirements of the \texttt{RunModel} interface (e.g., for multioutput models, generating a thin adapter that reshapes the model's output to the format expected by \texttt{GeneralizedSobolSensitivity}). 

	\subsubsection*{The Contextual Bandit Loop} Groups B, C, D iterates until convergence or budget exhaustion.

	\subsubsection*{Group B: Strategy Team (policy operator $\pi$)} At each iteration the \texttt{Strategist} receives the full history $\mathcal{H}_{n-1}$, the context $\mathbf{x}$, and the \texttt{Advisor}'s diagnosis from the previous round, and proposes the next action $A_n$: a method name together with its configuration parameters (e.g., $N_{\mathrm{samples}}$, PCE degree, Morris trajectories). The proposal is then submitted to the \texttt{Critic}, which analyzes it in two stages: first a deterministic rule-based check that enforces hard constraints at zero cost, then an LLM-based check for subtler issues such as redundancy with recent history or unjustified complexity increases. Only approved proposals advance. 
	
	\textit{Hyperparameter selection} (e.g., the number of samples, polynomial degree, or number of trajectories associated with the chosen estimator), is delegated to the \texttt{Advisor}. At the first iteration, when no prior observation is available, hyperparameters are initialized from the context: the minimum evaluation budget required by the estimator given the input dimensionality, scaled by a safety factor and limited at the available budget. From the second iteration onward, the \texttt{Advisor}'s \textit{diagnostic scheme} carries an explicit hyperparameter prescription extracted from its diagnosis (e.g., a recommended sample size or polynomial degree), which the \texttt{Strategist} adopts directly after validation by the Critic against the feasibility constraints of the action space. This design replaces blind exploration with informed recommendation: the \texttt{Advisor} reasons about the observed results to prescribe the configuration most likely to improve them, while the Critic ensures the prescription is physically feasible. The bandit then learns method quality, since the reward it observes already reflects the hyperparameter setting recommended by the Advisor.


\subsubsection*{Group C: Implementation Team (operator $\mathcal{I}$)} 
The \texttt{Study Agent} analyzes the retrieved template from the UQpy corpus $\mathcal{D}$ to supply a focused template, mitigating the ``Lost in the Middle" degradation caused by excessive context (where information embedded deep inside large prompts is often ignored or underutilized \cite{liu2023lost}), and produces a structured modification plan identifying exactly which code blocks of the template must be changed. The \texttt{Refactor Agent} then applies those modifications sequentially to the relevant code sections, receiving a cheat sheet inline to ground its code generation in correct library usage. This strategy, modifying targeted portions of the script rather than regenerating the entire program, reduces the autoregressive error accumulation inherent in long-context generation. The \texttt{Inspector} then performs a structural checklist audit of the assembled code $S_n$ against the strategy report, verifying that the correct UQpy class is implemented, parameters match, imports are present, distributions are configured, and outputs are saved. If the code crashes at runtime, the \texttt{Debugger} receives the full error trace and produces a structured diagnostic report identifying the root cause, the affected code sections, and targeted fix instructions. These instructions are passed explicitly to the \texttt{Refactor Agent}, which applies them to produce a corrected script that is immediately re-executed. This constitutes a closed self-healing loop, \texttt{Debugger} $\to$ \texttt{Refactor Agent} $\to$ \texttt{Execution Runner}.

	\subsubsection*{Group D: Execution and Evaluation (operators $\mathcal{E}$ and $\mathcal{A}$)} The \texttt{Execution Runner}, a pure Python function with no LLM call, writes $S_n$, runs it in a subprocess, and collects the numerical results, plots, and execution logs that constitute the observation $O_n$. The \texttt{Advisor} then evaluates $O_n$ and produces the scalar reward $R_n$ together with a diagnosis and prescribed correction for the next iteration. The \texttt{Advisor} is the only multimodal agent in the pipeline: it receives both the numerical results and the generated plots, allowing it to detect visual anomalies, such as non-converging index trajectories, that are invisible to purely numerical evaluation. The \texttt{Register} closes the loop by appending $(A_n, S_n, O_n, R_n)$ to the history $\mathcal{H}_n$ and verifying the submartingale condition $R_n \geq R_{n-1}$, completing one step of the online learning cycle.
	
	
	
	\begin{table}[ht]
		\centering
		\caption{Agent assignments, models, and primary outputs.}
		\label{tab:agents}
		\begin{tabular}{lllll}
			\toprule
			\textbf{Agent} & \textbf{Group} & \textbf{Operator} & \textbf{Model} & \textbf{Output} \\
			\midrule
			Coordinator      & A & -           & Claude Haiku        & $\mathbf{x}$, \texttt{ProblemScheme} \\
			Gatekeeper       & A & -           & Mistral 7B (local)  & routing, $\mathcal{M}_n$ \\
			Model Translator & A & -           & Claude Haiku        & \texttt{user\_model} ($f$) \\
			Strategist       & B       & $\pi$         & Claude Haiku        & $A_n$, \texttt{strategy\_report} \\
			Critic           & B       & $\pi$         & Mistral 7B (local)  & approved / rejected \\
			Study Agent      & C       & $\mathcal{I}$ & Claude Haiku        & \texttt{cell\_map} \\
			Refactor Agent   & C       & $\mathcal{I}$ & Claude Haiku        & $S_n$ \\
			Inspector        & C       & $\mathcal{I}$ & Mistral 7B (local)  & approved / rejected \\
			Debugger         & C       & $\mathcal{I}$ & Claude Haiku        & \texttt{fix\_instructions} \\
			Execution Runner & D       & $\mathcal{E}$ & Pure code           & $O_n$ \\
			Advisor          & D       & $\mathcal{A}$ & Claude Haiku  & $R_n$, diagnosis \\
			Register         & D       & -           & Pure code           & $\mathcal{H}_{n+1}$, \texttt{DiagnostiScheme} \\
			\bottomrule
		\end{tabular}
	\end{table}
	
	The learning intelligence of the system resides in two complementary structures. The first is the policy $\pi$, which accumulates cross-session evidence about which method configurations reliably produce high rewards for a given problem context; this is the statistical memory of the system. The second is the history $\mathcal{H}_n$, which records the full sequence $(A_i, S_i, O_i, R_i)_{i=1}^{n}$ and conditions every agent decision within a session. Together they implement a two-timescale learning architecture: $\mathcal{H}_n$ drives within-session adaptation through the \texttt{Strategist} and \texttt{Advisor}, while $\pi$ drives cross-session transfer through the bandit policy. All other components: the LLM calls, the checkpoints, the \texttt{Execution Runner}, are memoryless given these two structures: their outputs depend only on their immediate inputs.
	
	\subsection{Semantic Checkpoints}\label{sec:cp}
	
	\textcolor{black}{A fundamental assumption of the pipeline is that each agent operates autonomously: it receives an input and produces an output that becomes the input of the next agent. No agent has domain expertise in the task of its successor, and no output is manually engineered to match the precise information needs of the receiving agent. This autonomy is a design choice and it is what makes the system general and extensible, but it introduces a structural risk. When agents are free to formulate their own outputs, those outputs may be semantically coherent from the producing agent's perspective yet informationally inadequate from the receiving agent's perspective: a diagnosis that does not reference the numerical results it claims to interpret, a code plan that references the wrong estimator class, or a strategy report whose method name does not match the executable that was generated. These mismatches produce incorrect reward signals that corrupt the bandit's learning. Rather than constraining agent autonomy through manual output engineering, which would require anticipating every possible failure mode at design time, the framework mitigates this risk through the semantic checkpoint mechanism. Each checkpoint is placed at a specific agent boundary, compares the outputs of adjacent agents using a similarity measure, and either blocks the transition or passes a structured warning downstream when the comparison falls below a calibrated threshold. The checkpoints do not restrict what agents can produce; they verify that what was produced is coherent with what was expected, restoring the causal link between policy decision and reward signal without sacrificing agent autonomy. Some examples from the implementation of the framework illustrating the consequences of checkpoint ablation are provided in Section \ref{sec:experiments}.}
	
	A method selected by the \texttt{Strategist} may not survive intact through code generation, execution, and evaluation, not because any single agent fails catastrophically, but because small semantic distortions accumulate across transitions. This is the so-called \textit{semantic drift}. If the generated code $S_n$ does not faithfully implement the selected action $A_n$, then the reward $R_n$ measures the quality of the code, not the quality of the method, and the bandit loop learns the wrong thing.
	
The proposed multi-agent framework addresses this through a set of seven \textit{cosine-similarity checkpoints} (CP0--CP7), each placed at a specific agent boundary and targeting a distinct failure mode in the causal chain $A_n \to S_n \to O_n \to R_n$. Each checkpoint compares two adjacent pipeline outputs by first encoding them as semantic embedding vectors $\mathbf{u}$ and $\mathbf{v}$, where $\mathbf{u}$ represents the embedding of the upstream agent output and $\mathbf{v}$ the embedding of the downstream agent output being validated. Their semantic alignment is then quantified through cosine similarity:
\begin{equation}\label{eq:cosine}
\mathrm{sim}(\mathbf{u}, \mathbf{v}) =
\frac{\mathbf{u}\cdot\mathbf{v}}
{|\mathbf{u}||\mathbf{v}|}
\in [-1,1].
\end{equation}
In practice, the embeddings $\mathbf{u}$ and $\mathbf{v}$ are computed using the pretrained sentence-transformer \texttt{all-MiniLM-L6-v2} \cite{koronaki2025implementing} , which maps textual agent outputs into 384-dimensional normalized embedding vectors prior to cosine similarity evaluation.

	A transition is accepted when this similarity exceeds a threshold $\theta_{n}$ calibrated adaptively during the session: it tightens when the system consistently produces coherent transitions and relaxes when the problem is harder to match. Blocking checkpoints stop the pipeline and trigger a retry when they fail. This mechanism prevents distorted signals from propagating downstream. The non-blocking ones issue a structured warning and pass the mismatch as diagnostic information to the next agent. Table \ref{tab:checkpoints} lists all seven checkpoints with their compared quantities, blocking status, and failure actions, while Table \ref{tab:cp_params} provides the default threshold parameters.
	
	Prior to the bandit loop, a checkpoint CP0 compares the \textit{problem scheme} of the current session against an archive of completed sessions using a similarity score. CP0 is non-blocking: when a sufficiently similar past problem is identified, it warm-starts the contextual bandit by reducing its exploration, providing an informative prior that accelerates convergence without overriding the online learning mechanism. When no sufficiently similar problem is found in the archive, CP0 identifies the current session as anomalous, a problem instance that occupies a previously unseen region of the problem space, and responds by setting the exploration, ensuring that the bandit diversifies its method selections rather than defaulting to a single strategy without evidential support. This anomaly-detection role is a direct consequence of the similarity scoring: a low similarity score is not merely the absence of a warm start, it is a positive signal that the current problem is structurally novel, and the system responds to it as such by prioritizing variability over exploitation. Unlike CP1–CP7, which enforce semantic coherence within a session, CP0 enforces cross-session knowledge transfer and anomaly recognition. As the archive grows, the boundary between familiar and anomalous problems sharpens, and the system's ability to distinguish routine from novel contexts improves incrementally, making CP0 the primary mechanism through which the framework adapts its exploration strategy to the diversity of problems it has encountered.
	
	The role of each checkpoint is distinct and non-redundant. The key distinction is between checkpoints that enforce \textit{action propagation}, ensuring the selected method survives intact through code generation (CP1, CP2, CP4, CP5), and those that enforce \textit{signal integrity}, ensuring the reward measures actual method quality rather than code quality (CP6, CP7), with CP3 playing a supporting advisory role.
	
	\begin{itemize}
		\item CP1 (user message $\to$ context vector) verifies that the \texttt{Coordinator} correctly parsed the user's problem description into a structured context (see Section \ref{sec:system}, Group A). A failure here means the entire session operates on a misrepresented problem, (e.g., the wrong dimensionality, the wrong task type, or the wrong distribution family). It uses a low threshold because the user's natural language and the structured context vector are not similar even when semantically correct (See \cref{fig:cp1} in Appendix \ref{app:cp}).
		
		\item CP2 (action $\to$ context vector) is the most critical blocking  checkpoint. It verifies that the method proposed by the \texttt{Strategist} is coherent with the problem context. CP2 does not enforce a specific method, it enforces contextual compatibility. A \texttt{Sobol} proposal passes CP2 when the context encodes a scalar output, sufficient budget, and no correlation constraints. The same \texttt{Sobol} proposal fails CP2 when the context encodes an explicit request for moment-free estimation, correlated inputs, or vector-valued output. This distinction is essential: CP2 is not a method filter, it is a coherence gate between what the problem requires and what the policy proposes.
		
		\item CP3 (retrieved template $\to$ strategy) is the only non-blocking informational checkpoint. A low similarity score here does not indicate an error: it indicates that the corpus has no close template for the requested method. This information is passed as a structured warning to the \texttt{Study Agent}, which then generates code from the cheatsheet rather than adapting a retrieved template.
		
		\item CP4 (cell map $\to$ strategy) verifies that the \texttt{Study Agent}'s implementation plan correctly reflects the selected action before any code is written. This is a planning-level gate: if the cell map references the wrong UQpy class, the wrong attribute names, or the wrong parameter configuration, CP4 blocks and the \texttt{Study Agent} replans. Catching drift at the planning stage is strictly cheaper than catching it after code generation, since replanning requires only a short LLM call whereas code regeneration requires a full \texttt{Refactor Agent} pass.
		
		\item CP5 (assembled code $\to$ strategy) verifies coherence after code assembly but before execution. CP5 is non-blocking because code and strategy are stylistically dissimilar by construction, making cosine similarity a noisy signal at this boundary. Instead, CP5 operates alongside the \texttt{Inspector}'s structural checklist, which is blocking: the \texttt{Inspector} verifies that the correct UQpy class is implemented, the correct attributes are read, and all required outputs are saved. Together CP5 and the \texttt{Inspector} form a two-stage coherence gate around code generation (See \cref{fig:cp5} in Appendix \ref{app:cp}).
		
		\item CP6 (\texttt{Advisor} diagnosis $\to$ observations) verifies that the \texttt{Advisor}'s textual evaluation is grounded in the actual numerical results and plots. A hallucinated diagnosis would corrupt the reward signal and cause the bandit to learn the wrong thing. CP6 is non-blocking in practice because these two elements are dissimilar regardless of semantic alignment, but the \texttt{Advisor}'s system prompt enforces explicit grounding by requiring all diagnostic claims to reference specific numerical values from the observation.
		
		\item CP7 (new action $\to$ previous action, inverted) CP7 is the only inverted checkpoint: it passes on \textit{low} similarity rather than high, flagging when the proposed action is too similar to the previous one. It is non-blocking, rather than stopping the pipeline, it issues a structured novelty warning that the \texttt{Strategist} reads before finalizing its proposal. The hard enforcement of action diversity operates one level deeper, in the bandit policy itself, which maintains a set of previously selected estimators and explicitly down-weights repeated selections in the policy.
	\end{itemize}
	
	\begin{tikzpicture}[node distance=0.35cm and 0.28cm]
		\node[seclabel] (title) at (0,0){}; 
		
		\node[cpbox,
		fill=cGoldFill, draw=cGoldBord,
		below=0.5cm of title, xshift=-8.5cm]
		(cp0box) {
			\textcolor{cGoldText}{\textbf{CP0}}\\[2pt]
			\textcolor{cGoldBord}{\small ProblemScheme}\\
			\textcolor{cGoldBord}{\small vs archive}
		};
		\node[blocking, below=0.15cm of cp0box]{\textcolor{cMuted} {\small\textit{Non-blocking}}};
		
		\hspace{.1cm}
		\node[cpbox,
		fill=cGreenFill, draw=cGreenBord,
		right=0.25cm of cp0box]
		(cp1box) {
			\textcolor{cGreenText}{\textbf{CP1}}\\[2pt]
			\textcolor{cGreenBord}{\small User request}\\
			\textcolor{cGreenBord}{\small vs context}
		};
		\node[blocking, below=0.15cm of cp1box] {\small\textit{Blocking}};
		\hspace{-.15cm}
		\node[cpbox,
		fill=cPurpFill, draw=cPurpBord,
		right=0.25cm of cp1box]
		(cp2box) {
			\textcolor{cPurpText}{\textbf{CP2}}\\[2pt]
			\textcolor{cPurpBord}{\small Strategy action}\\
			\textcolor{cPurpBord}{\small vs context}
		};
		\node[blocking, below=0.15cm of cp2box] {\small\textit{Blocking}};
		\hspace{-.15cm}
		\node[cpbox,
		fill=cOranFill, draw=cOranBord,
		right=0.25cm of cp2box]
		(cp3box) {
			\textcolor{cOranText}{\textbf{CP3}}\\[2pt]
			\textcolor{cOranBord}{\small Template + mods}\\
			\textcolor{cOranBord}{\small vs strategy}
		};
		\node[blocking, below=0.15cm of cp3box] {\textcolor{cMuted}{\small\textit{Non-blocking}}};
		\hspace{-.15cm}
		\node[cpbox,
		fill=cOranFill, draw=cOranBord,
		right=0.25cm of cp3box]
		(cp45box) {
			\textcolor{cOranText}{\textbf{CP4/CP5}}\\[2pt]
			\textcolor{cOranBord}{\small Assembled code}\\
			\textcolor{cOranBord}{\small vs strategy}
		};
		\node[blocking, below=0.15cm of cp45box] {\small\textit{Blocking}};
		\hspace{-.15cm}
		\node[cpbox,
		fill=cBlueFill, draw=cBlueBord,
		right=0.25cm of cp45box]
		(cp6box) {
			\textcolor{cBlueText}{\textbf{CP6}}\\[2pt]
			\textcolor{cBlueBord}{\small Diagnosis text}\\
			\textcolor{cBlueBord}{\small vs observations}
		};
		\node[blocking, below=0.15cm of cp6box] {\textcolor{cMuted}{\small\textit{Non-blocking}}};
		\hspace{-.15cm}
		\node[cpbox,
		fill=cGrayFill, draw=cGrayBord,
		right=0.25cm of cp6box]
		(cp7box) {
			\textcolor{cGrayText}{\textbf{CP7}}\\[2pt]
			\textcolor{cGrayBord}{\small New action}\\
			\textcolor{cGrayBord}{\small vs prev.\ action}
		};
		\node[blocking, below=0.15cm of cp7box] {\textcolor{cMuted}{\small\textit{Warning only}}};
		
		
	\end{tikzpicture}

	Taken together, the checkpoints implement a four-layer verification of the causal chain:
	\begin{itemize}
		\item \textit{Transfer layer} (CP0): operates across sessions. Before the bandit loop begins, CP0 compares the current problem against an archive of completed sessions. On a close match, it warm-starts the policy, reducing unnecessary early exploration. On a novel problem, it identifies the current context as anomalous and maximizes the exploration coefficient, ensuring the system adapts rather than defaulting to previously rewarded strategies that may be inappropriate for the new problem structure.
		\item \textit{Selection layer} (CP1, CP2): ensures the policy operates on a correctly represented problem and proposes a contextually valid action. CP1 verifies that the Coordinator has faithfully parsed the user's description into the structured context; CP2 verifies that the proposed method is compatible with that context before any code is written.
		
		\item \textit{Implementation layer} (CP3, CP4, CP5, \texttt{Inspector}): ensures the selected action is faithfully translated into executable code, catching semantic drift at the planning stage before it propagates to code generation. CP3 advises the Study Agent when corpus coverage is low; CP4 verifies the implementation plan before any code is written; CP5 and the Inspector together audit the assembled code against the strategy report's required attributes and forbidden attributes derived from the method scheme.
		
		\item \textit{Evaluation layer} (CP6, CP7): ensures the reward signal measures actual method quality. CP6 verifies that the Advisor's diagnosis is grounded in the actual numerical results and plots, preventing hallucinated evaluations from corrupting the bandit signal. CP7 maintains the action diversity that makes the reward signal informative across iterations: by flagging insufficient novelty, it prevents the system from collapsing to a single repeated action, which would reduce the mutual information $I(\mathbf{A}; \mathbf{O} \mid \mathbf{x})$ to zero regardless of how reliably each individual transition is preserved.
	\end{itemize}

	This separation of concerns reflects the decomposition of system empowerment in \cref{def:system_empowerment}: the selection layer maximizes the statistical component by ensuring the policy acts on correct information; the implementation layer preserves the communicative component by ensuring that the action survives code generation intact; the evaluation layer closes the loop by ensuring that the reward signal is accurate and diverse enough to drive learning.
	
	\begin{table}[ht]
		\centering
		\caption{Semantic checkpoints: compared quantities, blocking flag, and failure action.}
		\label{tab:checkpoints}
		\begin{tabular}{clcl}
			\toprule
			\textbf{CP} & \textbf{Compared quantities} & \textbf{Blocking} & \textbf{On failure} \\
			\midrule
			CP0 & Context vector (current) vs context archive & No & Warm-start signals set to neutral; bandit explores. \\
			CP1 & User message vs context vector & Yes & Re-prompt Coordinator. \\
			CP2 & Action vs context vector & Yes & Critic rejects; Strategist retries. \\
			CP3 & Template vs strategy & No  & Warning; mismatch list to Study Agent. \\
			CP4 & Cell map vs strategy & Yes & Study Agent retries. \\
			CP5 & Assembled code vs strategy & Yes & Inspector rejects; Refactor retries. \\
			CP6 & Diagnosis vs observations & No & Warning; Advisor prompted for re-grounding. \\ 
			CP7 & New action vs previous action & No 
			& Warning passed to Strategist; pipeline continues.\\
			\bottomrule
		\end{tabular}
	\end{table}

	\subsubsection{Adaptive thresholding} 
	
	A single fixed threshold applied uniformly leads to two failure modes: too low and semantically incoherent transitions pass; too high and valid transitions are incorrectly rejected. The multi-agentic framework avoids both by treating each threshold as a running estimate of what a coherent transition looks like for that specific checkpoint, updated from the similarity scores observed in the current session. During the first iterations the system uses fixed hand-tuned values as a prior. Once sufficient observations are available, the threshold adapts: it secures when the system consistently produces coherent transitions and relaxes when the problem is harder to match. The running statistics are stored in the history by the \texttt{Register} and read by the checkpoint evaluator at the start of each iteration, closing a feedback loop in which past coherence levels inform future acceptance criteria. A threshold that decreases monotonically throughout a session is itself a diagnostic signal: it indicates that the system is becoming progressively more permissive, which typically reflects corpus coverage gaps or prompt quality issues rather than authentic improvement in pipeline coherence.
	
	The initial threshold values are not chosen arbitrarily, they are validated against the distribution of cosine similarities one would observe between completely unrelated text pairs drawn from the same domain. Additionally, a null distribution captures the baseline similarity that arises simply from two texts belonging to the same SA/UQ vocabulary, without any semantic relationship between them. A transition is accepted only if its similarity is high enough to be statistically distinguishable from such a random pairing at a chosen significance level. This reframes each checkpoint as a hypothesis test: the null hypothesis is that the two texts are unrelated, and a passing score is evidence against it. The adaptive floor prevents the threshold from drifting below the null-calibrated level during a session, ensuring that no matter how permissive the adaptive rule becomes, it never accepts transitions that are indistinguishable from random pairings.
	
	\begin{table}[ht]
		\centering
		\caption{Default adaptive threshold parameters per checkpoint.}
		\label{tab:cp_params}
		\begin{tabular}{llccc}
			\toprule
			\textbf{CP} & $\theta_0$ & $\theta_{\min}$ & \textbf{Adaptive} & \textbf{Type} \\
			\midrule
			CP0 & 0.70 & --- & No & Composite similarity score \\
			CP1 & 0.25 & 0.15 & Yes & Cosine (sentence embeddings) \\
			CP2 & 0.30 & 0.20 & Yes & Cosine (sentence embeddings) \\
			CP3 & 0.30 & 0.20 & Yes & Cosine (sentence embeddings) \\
			CP4 & 0.35 & 0.25 & Yes & Cosine (sentence embeddings) \\
			CP5 & 0.35 & 0.25 & Yes & Cosine (sentence embeddings) \\
			CP6 & 0.30 & 0.20 & Yes & Cosine (sentence embeddings) \\
			CP7 & 0.35 & 0.25 & Yes & Cosine, inverted \\
			\bottomrule
		\end{tabular}\\
	\end{table}

	\subsubsection{Checkpoints as empowerment stabilizers} 
	
	Taken together, the checkpoints implement the communicative component of system empowerment identified in \cref{def:system_empowerment}. Each blocking checkpoint is placed precisely at a transition where \textit{semantic drift} can cut the causal link $A_n \to S_n \to O_n \to R_n$, catching and retrying that transition before the distortion propagates downstream. 
	
	The semantic checkpoints are not just quality gates, they are the system's mechanism for maintaining agent empowerment across the pipeline. Empowerment, understood as the mutual information between an agent's actions and the future states it can reach, degrades whenever the agent loses coherence with its own outputs. Each CP measures exactly that: whether the current agent's output is consistent with what the previous agent requested. A failing CP does not stop the pipeline arbitrarily, it indicates that the agent is about to enter a region of the state space where its future actions will have diminished influence, because the code it generates, the template it retrieves, or the diagnosis it produces no longer reflects the problem it was given.
	
	CP7 complements this from a different direction: by rejecting consecutive iterations that select the same method, it prevents the system from collapsing to a single repeated action. A system that always selects the same method produces observations that are independent of its action choices, making $I(A_n; O_n \mid \mathbf{x}) = 0$ and reducing system empowerment to zero regardless of how reliably each individual transition is preserved. CP0 works in the opposite direction: by matching the current problem against historical archive entries, it transfers accumulated empowerment from past sessions, giving the agent a richer starting distribution over future states than it would have from a cold start.\\
	
	The schemes: ProblemScheme, MethodScheme, and DiagnosticScheme, are the structured representations of state that make empowerment measurable in the system. An agent can only exert influence over future states if it can represent them precisely enough to act on. Free-text passing between agents collapses that precision: a diagnosis written as a sentence cannot be acted on by a policy the way a DiagnosticScheme with a typed root cause, prescribed estimator, and warm hyperparams can. The schemes enforce that each agent transition carries not just a result, but a structured description of the reachable state space available to the next agent.
	
	MethodScheme in particular encodes empowerment directly through its budget status: before a single line of code is written, the system knows whether the chosen method can produce meaningful indices within the available budget, or whether it is heading into a region of guaranteed failure. DiagnosticScheme closes the loop across iterations it is the mechanism by which information about a past failure is converted into a structured prior that expands the action space of the next iteration's Strategist, rather than being lost as unstructured text.

	\section{Results}\label{sec:experiments}

	\subsection{Checkpoint Ablation (Example 1): Semantic Drift Under Adversarial Prompting}\label{sec:checkpoint_ablation1}
	
	To evaluate the communicative component of system empowerment (see Definition \ref{def:system_empowerment}), we designed an adversarial prompt that explicitly requests a moment-free estimator for a non-linear structural model:
	\begin{equation}
		f(X_1, X_2, X_3, X_4)
		\ = \
		X_1^3 + X_2 X_3 + e^{0.1 X_4} - X_1 X_4,
		\label{eq:structural_model}
	\end{equation}
	with $X_1 \sim \mathcal{U}(-2,2)$, $X_2 \sim \mathcal{U}(0,3)$, $X_3 \sim \mathcal{U}(1,4)$, $X_4 \sim \mathcal{U}(-1,1)$, and budget $N = 15,000$.
	The prompt contains the explicit constraint \emph{``do not use variance-based Sobol indices; I request a moment-free estimator such as Chatterjee or Cramér-von Mises''}.
	
	This creates a direct conflict with the dominant template in the UQpy corpus, which is Sobol-based, making silent substitution the path of least resistance for the Implementation Team.
	
	We ran three sessions under each condition. In the \emph{no-checkpoint} (no-CP) condition, all similarity thresholds were set to $\theta_i = -1$, so that every checkpoint (CP) passed unconditionally. In the \emph{checkpoint} condition, the system operated normally (Table \ref{tab:cp_params}).
	
	Table \ref{tab:ablation} reports, for each run, the method selected by the bandit policy, the method implemented in the generated code $S_n$, whether they match, the total reward $R_n$, and the number of blocking checkpoint events.
	
	\begin{table}[ht]
		\centering
		\caption{Action-code fidelity under checkpoint ablation. A mismatch means the reward $R_n$ is attributed to a method different from the one the bandit selected, corrupting the policy update.}
		\label{tab:ablation}
		\small
		\begin{tabular}{llllcc}
			\toprule
			\textbf{Condition} & \textbf{Bandit selected} & \textbf{Code ran}
			& \textbf{Match} & $R_n$ & \textbf{CP events} \\
			\midrule
			No-CP-1 & \texttt{Sobol}       & \texttt{Chatterjee} & \textcolor{red}{\ding{55}} & 82 & - \\
			No-CP-2 & \texttt{Sobol}  & \texttt{Sobol}      & \textcolor{green!60!black}{\ding{51}} & 74 & - \\
			No-CP-3 & \texttt{Sobol}  & \texttt{Morris} & \textcolor{red}{\ding{55}} & 61 & - \\
			\midrule
			CP 1    & \texttt{Sobol}  & \texttt{Sobol} & \textcolor{green!60!black}{\ding{51}} & 80 & \textit{CP2} $\times$1 \\
			CP 2    & \texttt{Sobol}         & \texttt{Sobol}        & \textcolor{green!60!black}{\ding{51}} & 70 & \textit{CP3} $\times$1             \\
			CP 3    & \texttt{Sobol}  & \texttt{Sobol} & \textcolor{green!60!black}{\ding{51}} & 53 & \textit{CP2} $\times$1 \\
			\bottomrule
		\end{tabular}
	\end{table}
	
	Without checkpoints, 2 of 3 runs produced action-code mismatches in which the reward signal was attributed to a method different from the one selected by the bandit, directly corrupting the policy update. With checkpoints active, we observed high fidelity; CP events prevented drift before code generation.
	
	Runs No-CP-1 and No-CP-3 are documented by the system's own diagnostic output. The policy selected pipeline \texttt{D2:Sobol}, yet the generated code implemented \texttt{ChatterjeeSensitivity} and \texttt{MorrisSensitivity}, producing results with $N=10,000$ samples. The \texttt{Advisor} correctly diagnozed the execution:
	\begin{quote}\small
		\textit{``Chatterjee indices successfully computed ($N=10,000$). $X_1$ dominates (0.318), $X_2$ secondary (0.213), $X_3$ weak (0.069), $X_4$ near-zero ($-0.002$).
			\textbf{Critical}: $X_4$ index is negative, violating Chatterjee non-negativity. The sum $= 0.599$ suggests that the effects of the interaction are not taken into account. Dependency flags warn $X_2$-$X_3$ coupling and $X_1$-$X_4$ interaction, but Chatterjee assumes independence.''}
	\end{quote}
	
	The \texttt{Advisor} further highlighted three scientific surprises: (i) the negative index $X_4$ ($-0.002$) violates Chatterjee monotonicity, attributable to the weak exponential term $e^{0.1 X_4}$ over $[-1,1]$ being dominated by the $-X_1 X_4$ interaction; (ii) $X_1$ dominates ($0.318$), confirming the cubic prior, but $X_4$ near-zero contradicts an expected exponential influence; (iii) the index sum $0.599 \ll 1$ despite the known $X_2$-$X_3$ multiplicative coupling, suggesting the rank-based Chatterjee metric misses interaction structure.
	
	Despite this accurate post-analysis, the \texttt{Advisor} diagnostic did not have an effect on the upstream action-code mismatch: the reward $R_n = 82$ was credited to the Sobol arm in the policy update, while measuring the quality of the Chatterjee method. This illustrates a fundamental asymmetry: the \texttt{Advisor}'s diagnostic capability is a \emph{necessary but not sufficient} condition for pipeline integrity. Checkpoints act \emph{before} execution to prevent drift; the \texttt{Advisor} detects drift \emph{after} execution but cannot change its effect on the learning signal.
	
	Figure \ref{fig:checkpoint_drift} contrasts the causal chains $A_n \to S_n \to O_n \to R_n$ for run No-CP-1 and CP-1. In the no-checkpoint case the link between policy selection and code implementation is interrupted silently; the reward measures Chatterjee method quality but updates Sobol weights, injecting noise into the policy. In the checkpoint case, CP2 obseved a \texttt{Sobol} proposal inconsistent with the moment-free constraint in the context vector, and the \texttt{Strategist} retried with \texttt{Sobol}; CP4 subsequently verified that the \texttt{Study Agent}'s cell map planned to implement \texttt{Sobol} before the code was written. The causal chain remained intact, and the reward correctly updated the Sobol arm.
	
	These results provide empirical support for the claim of Section \ref{sec:cp}: checkpoints are not quality filters applied after the fact, but \emph{causal integrity mechanisms} that preserve mutual information $I(A_n;\ O_n \mid \mathbf{x})$ by preventing the action chosen by the policy from being silently replaced downstream. Without them, the bandit loop learns noise rather than method quality, regardless of how well the \texttt{Advisor} diagnoses individual runs.
	
	\begin{figure}[ht]
		\centering
		\begin{tikzpicture}[
			node distance = 1.6cm and 2.2cm,
			box/.style  = {draw, rounded corners, minimum width=2.2cm,
				minimum height=0.7cm, font=\small, align=center},
			arr/.style  = {-stealth, thick},
			broken/.style = {-stealth, thick, red, dashed},
			]
			\node[box]                         (pol)  {Policy\\$\pi$};
			\node[box, right=of pol]           (imp)  {Implementation\\Team};
			\node[box, right=of imp]           (exec) {Execution\\Runner};
			\node[box, right=of exec]          (adv)  {Advisor\\$\mathcal{A}$};
			\node[box, below=0.5cm of pol]     (lbl1) {\texttt{D2:Sobol}};
			\node[box, below=0.5cm of imp]     (lbl2) {\texttt{Chatterjee}\\\texttt{Sensitivity}};
			\node[box, below=0.5cm of exec]    (lbl3) {$O_n$: \texttt{S1\_chatterjee}};
			\node[box, below=0.5cm of adv]     (lbl4) {$R_n{=}82$\\$\to$ Sobol arm};
			
			\draw[arr]    (pol)  -- (imp);
			\draw[broken] (imp)  -- node[above]{\textcolor{red}{\ding{55}}} (exec);
			\draw[arr]    (exec) -- (adv);
			
			\draw[arr] (pol)  -- (lbl1);
			\draw[arr] (imp)  -- (lbl2);
			\draw[arr] (exec) -- (lbl3);
			\draw[arr] (adv)  -- (lbl4);
			
			\node[above=0.15cm of pol, font=\bfseries\small] {No-checkpoint (run No-CP-1)};
		\end{tikzpicture}
		
		\begin{tikzpicture}[
			node distance = 1.6cm and 2.2cm,
			box/.style  = {draw, rounded corners, minimum width=2.2cm,
				minimum height=0.7cm, font=\small, align=center},
			cp/.style   = {draw, diamond, fill=yellow!30, font=\scriptsize,
				inner sep=1pt, align=center},
			arr/.style  = {-stealth, thick},
			ok/.style   = {-stealth, thick, green!60!black},
			lbl/.style  = {align=center, font=\small},
			]
			\node[box]                     (pol)  {Policy\\$\pi$};
			\node[cp,  right=0.9cm of pol] (cp2)  {{CP2}\\\ding{51}};
			\node[box, right=0.9cm of cp2] (imp)  {Implementation\\Team};
			\node[cp,  right=0.9cm of imp] (cp4)  {{CP4}\\\ding{51}};
			\node[box, right=0.9cm of cp4] (exec) {Execution\\Runner};
			\node[box, right=0.9cm  of exec]      (adv)  {Advisor\\$\mathcal{A}$};
			
			\node[lbl, below=0.5cm of pol]  (lbl1) {\texttt{Sobol}};
			\node[lbl, below=0.5cm of imp]  (lbl2) {\texttt{Sobol}\\\texttt{Sensitivity}};
			\node[lbl, below=0.5cm of exec] (lbl3) {$O_n$: \texttt{S1\_Sobol}};
			\node[lbl, below=0.5cm of adv]  (lbl4) {$R_n{=}79$\\$\to$ Sobol arm};
			
			\draw[ok]  (pol)  -- (cp2);
			\draw[ok]  (cp2)  -- (imp);
			\draw[ok]  (imp)  -- (cp4);
			\draw[ok]  (cp4)  -- (exec);
			\draw[ok]  (exec) -- (adv);
			
			\draw[arr] (pol)  -- (lbl1);
			\draw[arr] (imp)  -- (lbl2);
			\draw[arr] (exec) -- (lbl3);
			\draw[arr] (adv)  -- (lbl4);
			
			\node[above=0.15cm of pol, font=\bfseries\small]
			{Checkpoint-active (run CP-1)};
		\end{tikzpicture}
		
		\caption{Causal chain $A_n \ \to\  S_n \ \to\  O_n \ \to\  R_n$ for the no-checkpoint run No-CP 1 (top) and the checkpoint run CP 1 (bottom). In the no-checkpoint case the dashed red arrow marks the silent substitution of \texttt{Chatterjee} for the selected \texttt{Sobol}, causing the reward $R_n=82$ to be credited to the wrong bandit arm. In the checkpoint case CP2 blocks the inconsistent proposal and CP4 verifies the cell map before code is written, preserving the integrity of the learning signal.}
		\label{fig:checkpoint_drift}
	\end{figure}

    \textcolor{black}{The failure observed in this example is not a consequence of the policy selecting the wrong method, nor of the Critic approving an incoherent proposal. The Strategist correctly identified Sobol as the appropriate estimator and the Critic confirmed its contextual validity. The drift originated at the retrieval stage: the corpus returned a template whose structural similarity to the strategy was low, meaning it had been written for a different estimator, and the Study Agent adapted it without correcting the output schema. The result was a syntactically valid, successfully executing script that measured the wrong quantity, and a reward signal that the bandit could not distinguish from actually using the Sobol estimator and genuinely failing.}

    \textcolor{black}{These results highlight that the scaffolding approach of ATHENA and the checkpoint approach represent two complementary theories of semantic coherence in multi-agent pipelines. The scaffolding operates preventively: expert knowledge injected into agent prompts reduces the probability of incoherent outputs by construction, and is effective when domain constraints can be codified in advance. Checkpoints operate correctively: by measuring semantic distance between adjacent agent outputs after reasoning has occurred, they detect mismatches that were not anticipated at design time and trigger a retry before distortion propagates downstream. The ablation results in this section illustrate the distinction empirically, removing individual checkpoints does not produce catastrophic failures but introduces systematic, method-specific biases in the reward signal that degrade bandit convergence, because the source of the mismatch is structural incompatibility between agent output schemas rather than agent ignorance. The two mechanisms are therefore not competing solutions to the same problem but solutions to different layers of it: scaffolding prevents drift where constraints are known; checkpoints catch it where they are not.}\\

    \subsection{Adversarial Robustness and the Role of Semantic Checkpoints}
   
   To empirically evaluate the contribution of the semantic checkpoint CP5 to pipeline robustness, we designed a controlled adversarial experiment on the Sobol G-function benchmark ($d=8$, $a=[0,1,4.5,9,99,99,99,99]$, $N=15{,}000$). The experiment introduces a \emph{method-swap drift}: a targeted perturbation that replaces the method field in the strategy report from \texttt{SobolSensitivity} to \texttt{MorrisSensitivity} immediately after the Strategist produces it and before it reaches the Refactor Agent. This drift simulates a realistic failure mode in which the implementation layer receives an inconsistent instruction, one that is syntactically valid and causes no runtime exception, but is semantically wrong with respect to the task objective.
   
   Two conditions were compared under identical problem configuration and budget:
   \begin{itemize}
   	\item \textbf{Ablated} (\texttt{no\_cp5 + method\_swap}): CP5 is disabled; the drift-corrupted strategy report reaches the Refactor Agent unverified.
   	\item \textbf{Full} (\texttt{full + method\_swap}): the complete pipeline is active; CP5 compares the assembled code against the original strategy report before execution.
   \end{itemize}
   
   In both conditions, the policy selected \texttt{SobolSensitivity} as the optimal estimator for iteration~1, consistent with the analytic structure of the G-function and the warm-start prior from the context archive. The method-swap drift then replaced \texttt{SobolSensitivity} with \texttt{MorrisSensitivity} in the strategy report before it reached the Refactor Agent. The key distinction between the two conditions is therefore not which method the policy chose, but whether the pipeline had a mechanism to verify that the chosen method was actually what got implemented.
   
\begin{figure}[h!]
	   	\centering
	   	\includegraphics[width=0.75\textwidth]{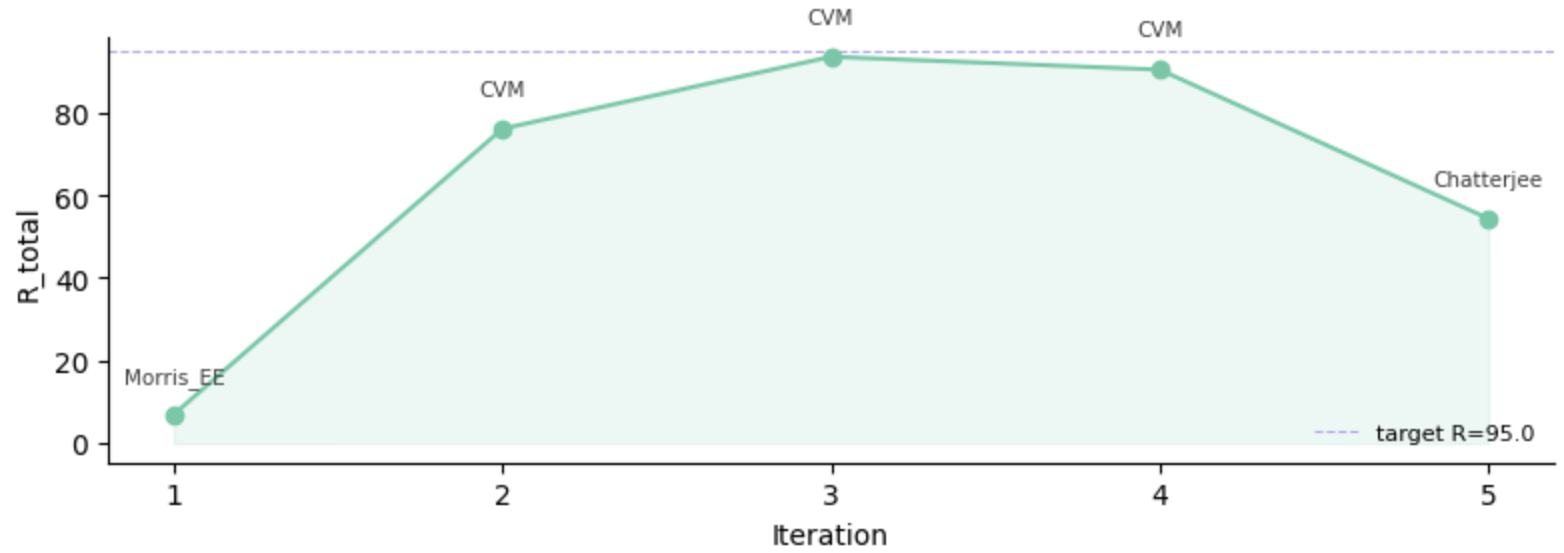}
	   	\caption{Reward trajectory under \texttt{no\_cp5 + method\_swap} ablation ($d=8$ Sobol G-function, $N=15{,}000$). Iteration~1 produces $R=7$ as the drift causes \texttt{MorrisSensitivity} code to execute in place of the requested Sobol estimator. The bandit recovers over iterations 2-4 via exploration of CVM, reaching $R=93.61$ at iteration 3. The dashed line marks the convergence threshold $R=85$. Under the full pipeline (\texttt{full + method\_swap}), CP5 prevents the drift from reaching execution and the system converges at $R=95.17$ in a single iteration.}
	   	\label{fig:cp5_ablation_trajectory}
   \end{figure}

   Under the ablated condition, this substitution went undetected: the Refactor Agent received the corrupted report and generated \texttt{MorrisSensitivity} code. The subprocess executed without error, because \texttt{MorrisSensitivity} is a valid \texttt{UQpy} operation. The Advisor correctly diagnosed the failure (\texttt{CRITICAL FAILURE: Morris screening indices reported instead of Sobol variance-based indices}), but the iteration was irrecoverable: $R_{\text{total}}=7$, $R_{\text{accuracy}}=0$. The system required three additional iterations of bandit exploration before converging at $R=93.61$ via the CVM estimator, which the policy selected after assigning a large penalty to the Sobol arm following the initial failure. This recovery trajectory is shown in Figure~\ref{fig:cp5_ablation_trajectory}, where the sharp drop at iteration 1 and the subsequent three-iteration recovery via bandit exploration are clearly visible.
   
   Under the full condition, CP5 compared the Morris-based assembled code against the \emph{original} strategy report, which still specified a variance-based sensitivity analysis, and detected the semantic mismatch. The similarity fell below the adaptive threshold $\theta_{\text{CP5}}$, triggering the Inspector, which rewrote the implementation using \texttt{PCESensitivity}, a method coherent with the task objective and with the distributions and budget available. The corrected code executed successfully on the first attempt, producing indices that matched the analytic reference to within $\text{MAE}(S_1) \approx 0.003$, and achieving $R_{\text{total}}=95.17$, an improvement of 88 reward points over the ablated condition in the same iteration.

   \begin{table}[h]
   	\centering
   	\caption{Comparison of ablated vs.\ full pipeline under method-swap drift.}
   	\label{tab:cp5_ablation}
   	\begin{tabular}{lcccc}
   		\hline
   		Condition & Method executed & $R_1$ & $R_{\text{accuracy},1}$ & Iterations to converge \\
   		\hline
   		\texttt{no\_cp5} & \texttt{MorrisSensitivity} (drift) & 7.0  & 0.0  & 3 (via CVM) \\
   		\texttt{full}    & \texttt{PCESensitivity} (corrected) & 95.17 & 43.67 & 1 \\
   		\hline
   	\end{tabular}
   \end{table}
   
   In the ablated condition, the method-swap drift effectively collapsed the action space available to the implementation agent: regardless of which estimator the policy selected, the corrupted strategy report caused the Refactor Agent to generate Morris code, reducing all downstream states to a single failure region. The agent's actions had lost causal influence over the outcome, the defining symptom of low empowerment.
   
   CP5 restores empowerment by acting as a \emph{coherence gate} at the implementation transition. By verifying that the assembled code is semantically consistent with the strategy report before committing to execution, CP5 preserves the space of reachable states: a passing CP5 confirms that the agent's action (code generation) will have the expected effect on the future state (numerical results). A failing CP5 signals that the causal chain has been broken, that the agent's action no longer maps to the expected outcome, and triggers a correction before the irreversible step of subprocess execution.
   
	\subsection{High-Dimensional Screening: Sobol Function with 15 Inputs}
	
	We evaluate the framework on a standard benchmark for high-dimensional sensitivity analysis, the Sobol G-function with 15 uncertain inputs:
	\begin{equation}
		f(X_1,\ldots,X_{15}) = \prod_{i=1}^{15}
		\frac{|4X_i - 2| + a_i}{1 + a_i},
		\qquad X_i \sim \mathcal{U}(0,1),
	\end{equation}
	
	where ${a} = [0,\ 0.5,\ 1,\ 2,\ 3,\ 4,\ 6,\ 8,\ 10,\ 15,\ 20,\ 30,\ 50,\ 80,\ 100]$. The coefficient $a_i$ controls the importance of input $X_i$: inputs with $a_i = 0$ are most influential and those with $a_i \geq 20$ are negligible. The task type is SA with budget $N = 50,000$ and the objective is screening, identifying which inputs produce more variability and which can be fixed, rather than precise index estimation.
	
	We ran three independent sessions of five iterations each, preserving the policy across sessions to test cross-session knowledge accumulation.
	Table \ref{tab:sobol_g_sessions} summarizes the method selected and the reward obtained in each iteration.
	
	\begin{table}[ht]
		\centering
		\caption{Estimator selections and rewards ($R_{\text{total}}$) across three independent sessions on the 15-dimensional Sobol function.}
		\label{tab:sobol_g_sessions}
		\small
		\begin{tabular}{cllr}
			\toprule
			\textbf{Session} & \textbf{Iteration} & \textbf{Estimator} & $R_{\text{total}}$ \\
			\midrule
			\multirow{5}{*}{1}
			& 1 & Sobol ($N=25,000$)   & \textbf{64.2} \\
			& 2 & PCESensitivity ($N=25,000$)         & 60.8 \\
			& 3 & ChatterjeeSensitivity ($N=25,000$)        & 62.8 \\
			& 4 & PCESensitivity ($N=50,000$)         & 62.8 \\
			& 5 & CVMSensitivity ($N=25,000$)               & 51.8 \\
			\midrule
			\multirow{5}{*}{2}
			& 1 & Sobol ($N=25,000$)   & 56.6 \\
			& 2 & MorrisSensitivity (200 traj.)            & 62.8 \\
			& 3 & PCESensitivity ($N=25,000$)         & \textbf{68.8} \\
			& 4 & PCESensitivity ($N=50,000$)         & 66.8 \\
			& 5 & CVMSensitivity ($N=25,000$)               & 66.3 \\
			\midrule
			\multirow{5}{*}{3}
			& 1 & ChatterjeeSensitivity ($N=25,000$)        & 60.8 \\
			& 2 & Sobol ($N=25,000$)   & 64.8 \\
			& 3 & Sobol ($N=50,000$)   & \textbf{75.5} \\
			& 4 & MorrisSensitivity (200 traj.)            & 62.8 \\
			& 5 & PCESensitivity ($N=25,000$)         & 49.8 \\
			\bottomrule
		\end{tabular}
	\end{table}
	
	Across the three sessions, the system explored five distinct estimators
	(\texttt{Sobol}, \texttt{PCESensitivity}, \texttt{ChatterjeeSensitivity}, \texttt{CVMSensitivity}, and \texttt{MorrisSensitivity}), visiting a different subset in each session. No single estimator was selected in all five iterations of any session. The set of estimators selected in Session 3 partially overlaps with Sessions 1 and 2, consistent with the policy concentrating probability mass on previously rewarded pipelines while retaining residual exploration.
	
	Despite the diversity of estimators and sample sizes, all methods that agreed on the same qualitative ordering: inputs $X_1$-$X_6$ (corresponding to $a_i \leq 4$) dominate the output variance, while $X_7$-$X_{15}$ ($a_i \geq 6$) are negligible for the screening objective. This finding is consistent with the analytic Sobol indices for this function, which assign first-order indices below $0.004$ to inputs with $a_i \geq 6$ (see Appendix \ref{sec:exp2} ).
	
	
	The three independent sessions on the 15-dimensional Sobol function provide direct evidence of cross-session knowledge accumulation. The best reward obtained in each session increased monotonically in all sessions: $R = 64.2$ in Session 1, $R = 68.8$ in Session 2, and $R = 75.5$ in Session 3, a cumulative gain of 11.3 points without any modification of the system architecture or prompt. This improvement is not attributable to luck in method selection: Session 3 selected \texttt{Sobol} with the full budget $N = 50,000$ in iteration 3, the same estimator used in iteration 1 of Session 1 but with $N = 25,000$, which had yielded $R = 64.2$. The policy, having observed across two prior sessions that \texttt{Sobol} with larger $N$ consistently outperformed Sobol with half the budget, concentrated the upper confidence bound on the full-budget variant and selected it earlier in the session horizon. This is precisely the mechanism that cross-session persistence is designed to produce: policy accumulated from prior sessions shift the exploitation-exploration balance toward pipelines that previously received high reward, so that the system arrives at its best configuration in fewer iterations each time it encounters a structurally similar problem.

	Three patterns emerged that are consistent with the known properties of the respective estimators. First, the Sobol indices with $N = 25,000$ produced negative $\hat{S}_i$ and $\hat{S}_{T_i}$ for the inputs $X_8$--$X_{15}$ in all three sessions. These constraint violations ($\hat{S}_{T_i} < 0$) were correctly flagged by the \texttt{Advisor} as estimation noise rather than structural failure, and prescribed increasing $N$ as cure. Second, PCE-Sobol with polynomial degree 4 produced valid indices with smooth monotonic decay when the sample size satisfied $N \geq 2.5 \times$ $n_\mathrm{poly}$ (for degree 4, $d_\mathrm{in}=15$). However, when degree 3 was used with insufficient basis orthogonality (Session 3, iteration 5), the decomposition collapsed: all 15 inputs received identical indices $\hat{S}_1 \approx 0.032$, a pathological result that the \texttt{Advisor} immediately identified from the uniform ranking as a numerical failure. Third, \texttt{CVMSensitivity} produced negative first-order indices in Session 1 but valid estimates in Session 2, suggesting sensitivity to random seed at this sample size; the Advisor flagged the Session 1 result as unreliable and did not recommend \texttt{CVMSensitivity} for the exploitation phase.
	
	In all sessions where \texttt{the Sobol} or \texttt{CVMSensitivity} estimates were made, the \texttt{Advisor} independently detected $\sum_i \hat{S}_{T_i} > 1$, which it correctly identified as evidence of significant higher-order interaction effects on the Sobol function. This finding is consistent with the known multiplicative structure of the model, which generates interaction terms that elevate $S_{T_i}$ above $S_{1,i}$ for every input. The Advisor reported this consistently as a scientific surprise relative to the expectation that a product-form model would exhibit additive behavior, a physically meaningful insight that emerged without any domain-specific prompting.

	\subsection{Anomaly detection via CP0} 
	
	To illustrate the anomaly-detection mechanism of the cross-session context checkpoint, we contrast two problems with the same task type but structurally different dimensionality and input distributions. The first is the cantilever beam sensitivity problem ($d_{\text{in}}=4$, Normal inputs, $N=20,000$), for which two completed sessions had been recorded in the CP0 archive. The second is a nonlinear thermal diffusion model ($d_{\text{in}}=20$, Uniform inputs, $N=100,000$), presented to the system for the first time with no prior archive entry.
	
	When the beam problem was submitted again, CP0 identified it as familiar by finding a close match in the archive. It responded by reducing the exploration, effectively instructing the bandit to exploit what it had already learned rather than explore a new method. The policy immediately selected the estimator that had performed best in previous sessions and achieved a high-quality result on the first iteration, without wasting any of the available budget on methods that the archive had already characterized as suboptimal.
	
	The thermal model produced the contrasting behavior that the anomaly-detection mechanism is designed to elicit. CP0 found no structurally similar problem in the archive: the similarity score fell below both match thresholds and classified the session as anomalous. Rather than silently defaulting to the same estimator that had worked for the beam, the system responded to the absence of prior knowledge in two ways. First, it increases exploration, marking that exploitation of archive knowledge was not justified and that the budget should be used to discover which methods are effective for this unseen problem class. Second, because the input dimensionality exceeded the screening threshold, the system activated a screening-first strategy, directing the Strategist to prefer methods that identify the active subspace before committing resources to a full variance decomposition. This is the qualitatively correct response: applying a Sobol estimator to a twenty-dimensional multiplicative model without prior screening risks exhausting the budget on a single estimator that may be poorly suited to the problem structure.
	
	The reward trajectory of the anomalous session reflected the exploration regime: the system tried multiple estimator families across iterations, rewards were more variable than in the familiar condition, and the cumulative regret grew approximately linearly over five iterations. This is the expected behavior for a novel problem on a short horizon, the system had not yet accumulated sufficient observations to identify the dominant estimator, but it was actively searching for one rather than repeating a strategy inherited from a structurally different problem. Crucially, at session end CP0 recorded the anomalous session in the archive, so that a future submission of a similar high-dimensional problem would begin with an informative prior rather than as a cold start.
	
	Together, these two conditions illustrate the dual role of CP0 in the empowerment framework. On familiar problems, it preserves and transfers the statistical knowledge accumulated across sessions, reducing the effective horizon needed to reach convergence. On anomalous problems, it recognizes the boundaries of that knowledge and responds with a principled exploration strategy calibrated to the structural novelty of the problem, preventing the system from applying learned heuristics in a regime where they have not been validated.
	
	Table \ref{tab:cp0_contrast} summarizes the contrast between the two conditions.
	
	\begin{table}[ht]
		\centering
		\caption{CP0 behaviour on a familiar vs.\ anomalous problem.}
		\label{tab:cp0_contrast}
		\begin{tabular}{lcc}
			\toprule
			& Cantilever beam     & Thermal diffusion   \\
			\midrule
			$d_{\text{in}}$          & 4                   & 20                  \\
			Input distribution       & Normal              & Uniform             \\
			$N$                      & 20{,}000            & 100{,}000           \\
			Archive entries (same task) & 2               & 0                   \\
			CP0 similarity           & 0.991               & 0.243               \\
			Match type               & close               & none (anomalous)    \\
			Screening-first          & No                  & Yes ($d\geq8$)      \\
			Method at iteration 1    & \texttt{Sobol}               & \texttt{Morris}        \\
			$R_1$                    & 91.0                & 72.0                \\
			\bottomrule
		\end{tabular}
	\end{table}

	\subsection{Permutation Robustness of the Learned Policy}
	
	A bandit policy that updates online accumulates evidence sequentially, raising the question of whether the final learned policy is a robust function of the problem context or a path-dependent artifact of the particular sequence of early explorations. If two sessions on the same problem, starting with different initial method selections, converge to substantially different distributions, the policy is brittle: what the system learns depends more on which methods happened to be tried first than on the underlying structure of the problem.
	
	The most direct way to measure this is to fix the first iteration of each session to a specific method and observe how the policy evolves from that constrained starting point across multiple runs. By systematically varying which method is forced in the first iteration: running one set of sessions starting from Sobol, another starting from Chatterjee, another from Morris, and comparing the reward trajectories and final distributions, it becomes possible to quantify how strongly the identity of the first exploration shapes the policy that eventually emerges. The path-dependence score, computed as the normalized variance of best rewards across these forced-start conditions, summarizes the result in a single interpretable number. A low score indicates that the bandit recovers from any starting point and converges to a consistent policy; a high score indicates that the first method tried acts as an anchor that the policy does not fully escape within the session horizon. This has direct implications for the design of the exploration schedule: if path dependence is high, the warm-start mechanism of CP0 becomes essential not just for accelerating convergence but for reducing the variance introduced by the accident of early exploration, since seeding the policy from a similar past session effectively replaces the random first step with an informed one. If path dependence is low, the system is already robust and the value of CP0 lies primarily in its efficiency gains rather than in its stabilizing effect on the learned policy. Investigating this relationship systematically: across problem types, dimensionalities, and archive sizes, constitutes a natural direction for future work, and the experimental infrastructure required to conduct such a study in a controlled and reproducible manner.

	\section{Conclusion}\label{sec:conclusion}
	
	We have presented a multi-agent framework for automated sensitivity analysis and uncertainty quantification that formulates method selection as an online learning problem. The system combines a contextual bandit loop with a hierarchy of specialized LLM agents, structured inter-agent communication, and a UQpy-grounded implementation layer, resulting in a pipeline that is transparent, debuggable, and extensible. A self-healing execution cycle further enhances robustness by resolving runtime errors through an explicit Debugger–Refactor–Execution loop.

A central contribution of this work is the interpretation of the multi-agent architecture through the lens of \textit{empowerment maximization} \citep{klyubin2005empowerment, yiu2025empowerment}. Empowerment, defined as the mutual information between actions and outcomes, provides a unifying principle that clarifies the requirements for reliable learning in such systems. In this view, effective automation requires both the ability to identify high-quality actions and guarantees that these actions are faithfully realized and evaluated.

The \textit{contextual bandit loop} implements the statistical component of empowerment by learning which method configurations causally determine high-quality outcomes for a given problem context. Its exploration–exploitation mechanism promotes variability, by probing under-explored actions, and controllability, by concentrating resources on reliable strategies. This learning extends across sessions through a similarity-based warm-starting mechanism: previously solved problems inform new ones, accelerating convergence in familiar regimes while preserving flexibility.

Complementing this, the \textit{semantic checkpoints} and structured schemes implement the communicative component of empowerment. They enforce consistency at each agent transition and prevent semantic drift, ensuring that the causal chain $A_n \rightarrow S_n \rightarrow O_n \rightarrow R_n$ remains intact. Without such guarantees, the reward signal becomes decoupled from the selected action, and the learning process degrades. From a broader perspective, these checkpoints can be interpreted as enforcing consistency across multiple representations of the same computational process. This viewpoint is closely related to ideas from manifold learning, where intrinsic structure is recovered by aligning different observations or modalities of a system \citep{
sroczynski2024learning}. In particular, multi-view extensions such as alternating diffusion \citep{lederman2018learning} recover common latent structure by enforcing agreement across heterogeneous measurements. In this sense, semantic checkpoints act as discrete analogues of such geometric consistency constraints, aligning policy decisions, generated code, execution outputs, and evaluations along a shared latent structure.

The \textit{cross-session context checkpoint} further extends this framework by enabling anomaly detection and adaptive exploration. By comparing the current problem against an archive of prior sessions, the system distinguishes between familiar and structurally novel contexts. For familiar problems, it warm-starts the policy, improving sample efficiency and convergence speed. For novel problems, it increases exploration and prioritizes variability, enabling the discovery of effective strategies in previously unseen regimes.

\textcolor{black}{Looking forward, several directions emerge. First, while cosine similarity provides a practical proxy for semantic consistency, more principled alternatives may improve robustness. One promising direction is the use of representation alignment methods inspired by manifold learning and multi-view data fusion, where consistency is enforced across heterogeneous observations of the same underlying process \citep{coifman2006diffusion, coifman2008graph, singer2009nonlinear, talmon2013empirical, lederman2018learning}. In particular, approaches based on diffusion geometry, alternating diffusion, and manifold alignment recover common latent structure by identifying representations that remain invariant across modalities, sensors, or partial observations \citep{lederman2015learning, ledoux2018alternating, sroczynski2024learning}. Closely related ideas also arise in equation-free and coarse-grained modeling frameworks, where consistency between microscopic simulations and macroscopic observables is enforced through lifting and restriction operators acting across scales \citep{kevrekidis2003equationfree, kevrekidis2004equationfree, kevrekidis2009equationfree}. From this perspective, semantic checkpoints may be viewed as discrete consistency operators acting across multiple representations of the same computational workflow. Additional robustness may also come from program equivalence and formal verification techniques \citep{leroy2009formal, clarke2018handbook}, as well as causal consistency tests based on intervention-response behavior \citep{pearl2009causality, peters2017elements}. More broadly, the empowerment perspective suggests deeper connections to causal representation learning and geometric organization of scientific workflows. This direction naturally connects with recent developments such as GRAFT-ATHENA, where scientific workflows are embedded into persistent graph-structured problem and method spaces supporting continual accumulation and transfer of computational experience. A promising future direction is the integration of these perspectives: combining GRAFT-ATHENA’s geometric memory and cross-problem methodological transfer with empowerment-aware semantic checkpoints could yield multi-agent systems that not only accumulate scientific experience across sessions, but also preserve semantic integrity and causal attribution within each execution trajectory.}

Overall, this work suggests a general design principle for scientific multi-agent systems: \textit{adaptive decision-making must be coupled with guarantees of faithful execution and evaluation}. Contextual bandits expand the set of controllable action–outcome relationships, while structured communication and consistency constraints preserve their integrity. This combination provides a principled foundation for robust automation of scientific workflows, and we anticipate that these ideas will extend naturally to broader domains in computational science and engineering.

	\bibliographystyle{plainnat}
	\bibliography{references}
	
	\newpage
	
\appendix

\section{Implementation Details}\label{sec:details}
	
	\subsection{Action Space and Constraint Filtering}
	
	The action space $\mathcal{M}$ is partitioned by task type. Rather than a flat set of methods, each action $A \in \mathcal{M}$ is a vector $A = (d_1, \ldots, d_7)$ specifying a complete analytical pipeline.
	
	For instance, the SA estimator dimension $\mathcal{D}_2$ covers the principal families of global sensitivity methods available in UQpy:
	$$\mathcal{D}_2^{\mathrm{SA}} = \{\texttt{Sobol},\ \texttt{ChatterjeeSen.},\ \texttt{CVMSen.},\ \texttt{MorrisSen.},\ \texttt{PCESen.},\ \texttt{Generalized\_SobolSen.}\}.$$ 
	
	Each method addresses a distinct class of problems.
	\begin{itemize}
		\item{\texttt{Sobol}} computes first-order and total-order variance-based indices \citep{sobol1993sensitivity} via Saltelli's estimator scheme, requiring $\mathcal{O}(N(d_{\mathrm{in}}+2))$ model evaluations \citep{saltelli2010variance}. It is the default choice for low-to-moderate dimensional problems with sufficient budget. \item{\texttt{Chatterjee}} computes the rank correlation coefficient of \citet{chatterjee2021new}, a distribution-free alternative that is consistent and computationally efficient for large $N$ but provides no interaction terms. 
		\item{\texttt{MorrisSensitivity}} applies the Morris elementary effects method \citep{morris1991factorial}, which ranks inputs at cost $\mathcal{O}(N(d_{\mathrm{in}}+1))$ and is the appropriate choice when $d_{\mathrm{in}} \geq 10$ and a full variance decomposition is computationally prohibitive. 
		\item{\texttt{PCESensitivity}} constructs a sparse polynomial chaos surrogate \citep{sudret2008global} and derives Sobol indices analytically from the expansion coefficients, achieving high accuracy at low model evaluation cost when the response is smooth and the budget satisfies $N \geq 500\ d_{\mathrm{in}}$. \item{\texttt{Generalized\_Sobol}} computes generalized Sobol indices \citep{gamboa2014sensitivity} for vector-valued models ($d_{\mathrm{out}} > 1$), aggregating variance contributions across output dimensions into scalar sensitivity measures via the \texttt{GeneralizedSobolSensitivity} class in UQpy.
	\end{itemize}

	\subsection*{Feasibility filtering}  Not all methods in $\mathcal{M}$ are applicable to every problem. At each iteration, the system computes the feasible subset by evaluating hard constraints derived from the context vector and removing methods that violate them. Three constraints are currently enforced and presented in \cref{feasible}.
	
	\begin{table}[ht]
		\centering
		\caption{Feasibility predicates for SA methods. A method is blocked when its predicate evaluates to False.}\label{feasible}
		\begin{tabular}{lll}
			\toprule
			\textbf{Method} & \textbf{Predicate $c_k(\mathbf{x})$} & \textbf{Rationale} \\
			\midrule
			Morris\_screening\_high\_dim & $d_{\mathrm{in}} \geq 10$ & Screening interpretation unreliable for low $d_\text{in}$ \\
			Gen\_Sobol\_multioutput & $d_{\mathrm{out}} > 1$ & Requires vector-valued output \\
			PCE\_based\_Sobol & $N \geq 500\ d_{\mathrm{in}}$ & Surrogate accuracy requires sufficient samples \\
			\bottomrule
		\end{tabular}
	\end{table}
	
	
	The remaining dimensions $\mathcal{D}_1, \mathcal{D}_3, \ldots, \mathcal{D}_7$ specify the sampling strategy, surrogate model, budget allocation, output aggregation, confidence intervals, and screening pre-filter respectively. The UQ action space $\mathcal{M}_{\mathrm{UQ}}$ follows the same vectorial structure, with $\mathcal{D}_1$ covering uncertainty propagation methods including \texttt{Monte Carlo sampling}, \texttt{Latin hypercube}, \texttt{importance sampling}, and \texttt{MCMC strategies}, \texttt{polynomial chaos expansion}, \texttt{Gaussian process}, and \texttt{MCMC-based Bayesian inference}.

\newpage
\section{Scheme: Structure and Inter-agent Flow}
\label{app:scheme}

The examples below are drawn from the beam sensitivity analysis described in Section \ref{sec:experiments}: a four-input multiplicative model $\delta(\mathbf{X}) = (P L^3)/(3EI)$ with Normal inputs and budget $N = 20,000$.

The three scheme types that structure inter-agent communication in the multi-agent system are described in Section \ref{sec:design}.  All three objects are passed as structured arguments between agents; no free text is exchanged at these boundaries.

\subsection{ProblemScheme}

The \texttt{ProblemScheme}, produced once by the \texttt{Coordinator} from the user's context vector, encodes the complete structural characterization of the problem. It is passed unchanged to multiple downstream components: CP0 (for archive comparison), the bandit policy (to condition similarity search and determine feasible actions), the \texttt{MethodScheme} builder, the Strategist (to constrain proposals), and the Critic (to validate feasibility).

Its primary function is twofold: first, it pre-computes the feasible estimator set $\mathcal{M}_n \subseteq \mathcal{M}$ by evaluating hard constraints (e.g., $N \geq 500(d_{\mathrm{in}}+2)$ for Sobol); second, it provides the feature vector used by CP0 for archive matching. It is computed once and held constant throughout the session.

\begin{lstlisting}[language=Python, caption={\texttt{ProblemScheme} example for the cantilever beam SA problem. Produced by the Coordinator before the bandit loop begins.}]
ProblemScheme(
 # Problem dimensions
 d_in          = 4,
 d_out         = 1,
 N_budget      = 20000,
 task_type     = "SA",
 
 # Structural classification
 output_class  = "scalar",
 model_class   = "multiplicative",
 dist_family   = ["Normal", "Normal", "Normal", "Normal"],
 
 # Input structure flags
 has_dependence       = False,  # independent inputs
 limit_state_defined  = False,  # no failure surface
 low_fidelity_model   = False,
 target_pdf_known     = False,
 
 # Dimensionality regime flags
 high_d_in_flag  = False,  # d_in < 20
 multi_out_flag  = False,  # d_out = 1
 field_out_flag  = False,
 
 # Pre-computed feasible estimator sets
 feasible_sa_estimators = ["Sobol", "Chatterjee", "CVM", "PCE_SA"],
 feasible_uq_estimators = ["MCS_moments", "PCE_moments"],
 
  # Pre-computed feasibility flags
 sobol_feasible  = True,   # N=20000 >= 500*(4+2)=3000
 pce_feasible    = True,   # N >= 2*C(d+3,3) = 70
 morris_feasible = False,  # d_in=4 < 6 (screening threshold)
)
\end{lstlisting}

\noindent The feasibility flags have direct downstream effects. For example, 
\begin{itemize}
	\item \texttt{morris\_feasible = False} blocks \texttt{Morris} from appearing in
	$\mathcal{M}_n$, preventing the Strategist from proposing it. 
	\item \texttt{high\_d\_in\_flag} = \texttt{True} and CP0 set \texttt{screening\_first = True}, directing the policy toward \texttt{Morris} on the first iteration.
\end{itemize}

\subsection{MethodScheme}

The \texttt{MethodScheme}, produced by the \texttt{BanditPolicy} and attached to each \texttt{PolicyDecision}, is built from the selected action and the \texttt{ProblemScheme}. It is consumed by three Implementation Team components: the Strategist (which reads \texttt{valid\_when}, \texttt{invalid\_when}, \texttt{n\_min\_value}, and \texttt{hyperparams} to configure method parameters without consulting the cheatsheet), the Critic (which reads \texttt{required\_attributes} and \texttt{forbidden\_attributes} to verify code correctness before execution), and the Inspector (which uses the same attribute lists for structural validation).

The \texttt{MethodScheme} serves as the primary mechanism through which the Implementation Team receives structured API knowledge rather than free-text prompts. It specifies the exact UQpy class to instantiate, the minimum budget derived from a closed-form formula, the attributes the generated code must read after execution, and the conditions under which the method is valid or invalid. By injecting this block directly into agent prompts, the framework eliminates the need for those agents to re-derive method requirements from a generic cheatsheet, ensuring consistency and reducing hallucination.

\begin{lstlisting}[language=Python, caption={MethodScheme for Sobol sensitivity analysis on the cantilever beam. Produced by BanditPolicy and attached to the PolicyDecision passed to the Strategist.}]
MethodScheme(
 # Action identification
 action_tuple   = SA_ActionTuple(D1="Sobol", D2="None",
 D3="Fixed_N", D4="Scalar"),
 task           = "SA",
 estimator      = "Sobol",
 index_type     = "variance-based",
 
 # What the estimator produces
 sobol_produces = ["S1", "ST"],
 sampling_scheme = "pick-freeze",
 
 # Budget requirements
 n_min_formula  = "500 * (d_in + 2)",
 n_min_value    = 3000,          # = 500 * (4 + 2)
 n_cost_actual  = 51000,         # = 8500 * (2*4 + 2), Saltelli
 budget_status  = "sufficient",  # N=20000 << n_cost_actual
 # actual cost uses n_samples=8500
 
 # UQpy API specification
 library_class     = "UQpy.sensitivity.SobolSensitivity",
 surrogate_class   = None,
 output_strategy   = "scalar",
 expected_output_shape = "(4,)",
 
 # Inspector checklist --- code MUST read these attributes
 required_attributes  = ["first_order_indices",
 "total_order_indices"],
 forbidden_attributes = [],
 
 # Level-2 bandit hyperparams for this iteration
 hyperparams = {"n_samples": 8500},
 
 # Validity conditions injected into Strategist prompt
 valid_when   = ["inputs statistically independent",
 "d_out == 1",
 "N >= 500 * (d_in + 2)"],
 invalid_when = ["correlated inputs (sum(ST) > 1 diagnostic)",
 "d_in > 30 without prior screening"],
)
\end{lstlisting}

\noindent The \texttt{required\_attributes} field is the direct input to the Inspector's structural checklist. When the generated code does not read \texttt{first\_order\_indices} from the UQpy sensitivity object, the Inspector rejects the code and the Refactor Agent is invoked: without this field, the Inspector would need to infer the correct attribute name from general knowledge, a frequent source of hallucinated API calls. The \texttt{budget\_status} field (here \texttt{sufficient}) feeds into \texttt{R\_optimality}: if the actual cost significantly exceeds the minimum, the Advisor penalises the iteration for inefficient use of the budget.

\subsection{DiagnosticScheme}

The \texttt{DiagnosticScheme}, produced by the \texttt{Advisor} at the end of each iteration through deterministic pattern matching on its free-text diagnosis, is consumed by three components. The Strategist at iteration $n+1$ reads \texttt{prescribed\_estimator}, \texttt{prescribed\_N\_factor}, and \texttt{prescribed\_hyperparam} to configure the next action. The Register reads \texttt{penalize\_action} and \texttt{block\_action} to update the bandit reward signal. The BanditPolicy Level-2 injects \texttt{prescribed\_hyperparam} as an observed arm in the bandit for the relevant estimator.

The \texttt{DiagnosticScheme} serves as the mechanism through which the Advisor's scientific judgment propagates into subsequent iterations without requiring the Strategist to parse natural language. It converts free-text diagnosis into a structured scheme with controlled-vocabulary fields: the \texttt{root\_cause} field determines the primary failure mode, the prescribed fields carry actionable corrections, and the penalize action and block action flags directly modify the reward passed to the policy. This structured approach ensures that diagnostic insights reliably inform both method selection and hyperparameter tuning in the next iteration.

\begin{lstlisting}[language=Python, caption={DiagnosticScheme produced by the Advisor after iteration 1 of the cantilever beam session. The Sobol estimator ran with N=8500 samples; sum(S1)=1.044 slightly exceeded 1 and confidence intervals were wide.}]
DiagnosticScheme(
 # Convergence assessment
 convergence_status = "partial",  # R=72.0, below threshold 85
 bottleneck_dim     = "R_accuracy",  # main reward gap is here
 reward             = 72.0,
 
 # Root cause (controlled vocabulary, derived by pattern matching
 # from Advisor free-text: "N=8500 insufficient, sum(S1)=1.044")
 root_cause = "insufficient_N",
 
 # Prescriptions for next iteration
 prescribed_estimator   = "Sobol",     # keep same estimator
 prescribed_surrogate   = None,        # no surrogate needed
 prescribed_output_d4   = None,        # scalar output unchanged
 prescribed_N_factor    = 2.0,         # double N: 8500 -> 17000
 prescribed_hyperparam  = {"n_samples": 17000},
 
 # Bandit signals
 penalize_action = False,  # partial result, not a failure
 block_action    = False,  # estimator is valid, do not block
 
 # Physical interpretation (free text, informational only)
 physical_insight = (
 "X2 (length) contributes only 4.5% despite cubic exponent; "
 "inverse nonlinearities in E and I dominate at nominal scales"
 ),
)
\end{lstlisting}

\noindent At iteration $n+2$, the Strategist reads \texttt{prescribed\_N\_factor = 2.0} and calls \texttt{compute\_method\_params} with \texttt{l2\_hyperparams = \{"n\_samples": 17000\}}. The BanditPolicy simultaneously records $({\tt n\_samples}=17000,\ R=72.0)$ as a new observation for the arm of Sobol, so that future sessions on structurally similar problems start with an informed prior on the appropriate sample size. If \texttt{block\_action} were \texttt{True} (as it would be for an \texttt{attribute\_error} root cause), the reward passed to policy for this iteration would be set to zero regardless of what the Advisor scored, preventing the policy from reinforcing a method that crashed.

Table \ref{tab:scheme_summary} summarizes the scope, producer, and consumers of each scheme type together with the pipeline boundary at which each transfer occurs.

\begin{table}[ht]
	\centering
	\caption{Scheme types: production scope, producing agent, consuming agents, and pipeline boundary.}
	\label{tab:scheme_summary}
	\small
	\begin{tabular}{llllll}
		\toprule
		Scheme & Scope & Produced by & Consumed by & Boundary \\
		\midrule
		ProblemScheme
		& once per session
		& Coordinator
		& CP0, BanditPolicy, Strategist
		& Phase 0 $\to$ Phase 1 \\[4pt]
		MethodScheme
		& once per iteration
		& BanditPolicy
		& Strategist, Refactor Agent, Inspector
		& Phase 1 $\to$ Phase 2 \\[4pt]
		DiagnosticScheme
		& once per iteration
		& Advisor
		& Register, BanditPolicy, Strategist
		& Phase 3 $\to$ Phase 1$_{n+1}$ \\
		\bottomrule
	\end{tabular}
\end{table}

\noindent
Together, the three schemes implement the communicative component of system empowerment identified in Definition \ref{def:system_empowerment}: each boundary where a scheme is transferred is a boundary where semantic drift could otherwise break the causal chain $A_n \to S_n \to O_n \to R_n$. The \texttt{ProblemScheme} ensures the policy operates on a correctly represented problem; the \texttt{MethodScheme} ensures the selected action survives intact through code generation; the \texttt{DiagnosticScheme} ensures the reward signal propagates back to the policy in a form that the bandit can directly act on.

\section{Examples of Inter-Agent Semantic Checkpoints}\label{app:cp}

    \begin{figure}[h!]
	   	\centering
	   	\includegraphics[width=0.75\textwidth]{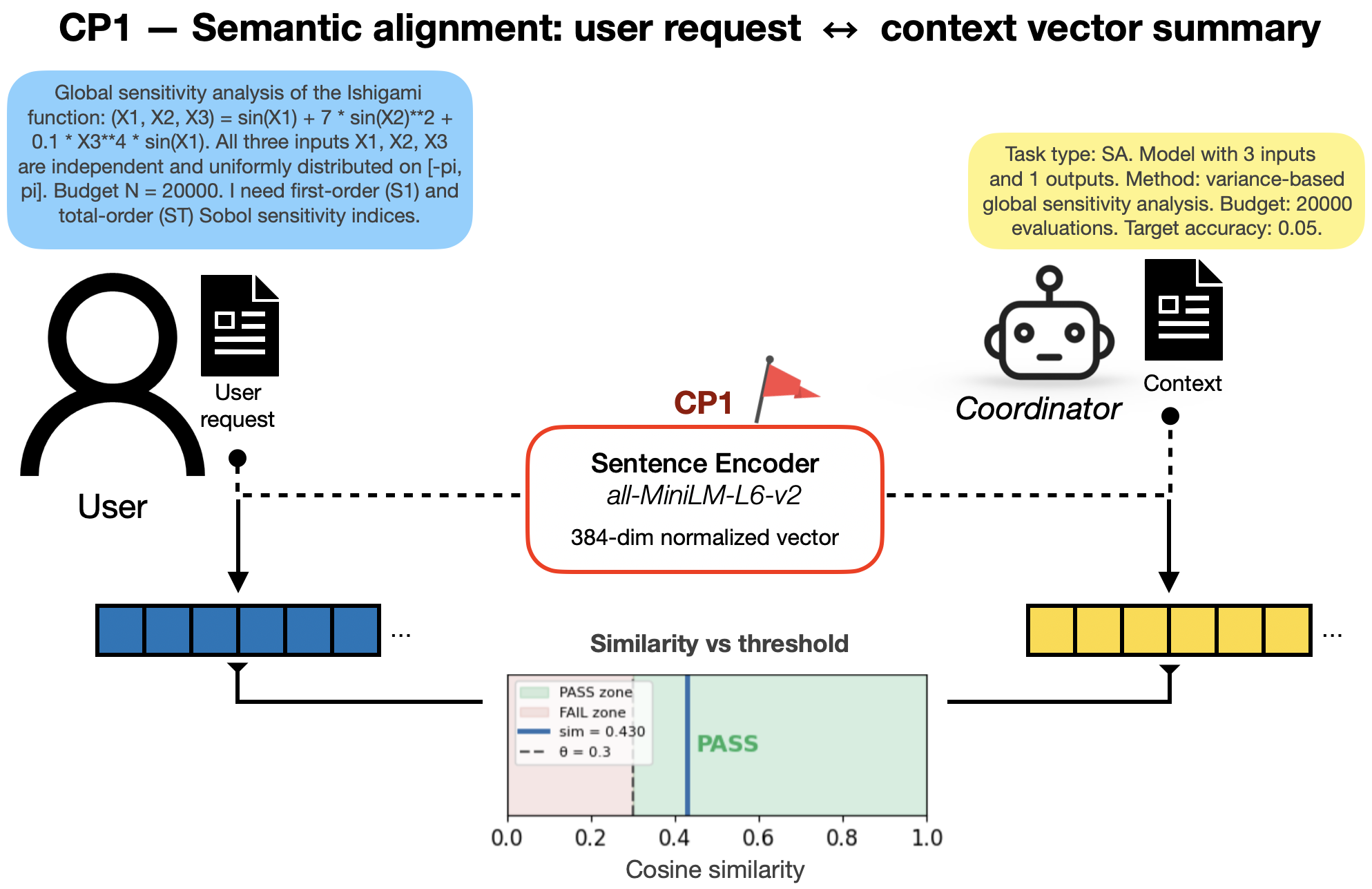}
	   	\caption{Semantic alignment between the raw user request (User output) and the structured context vector (Coordinator output), measured via cosine similarity of sentence embeddings.}
	   	\label{fig:cp1}
   \end{figure}

   \begin{figure}[h!]
	   	\centering
	   	\includegraphics[width=0.75\textwidth]{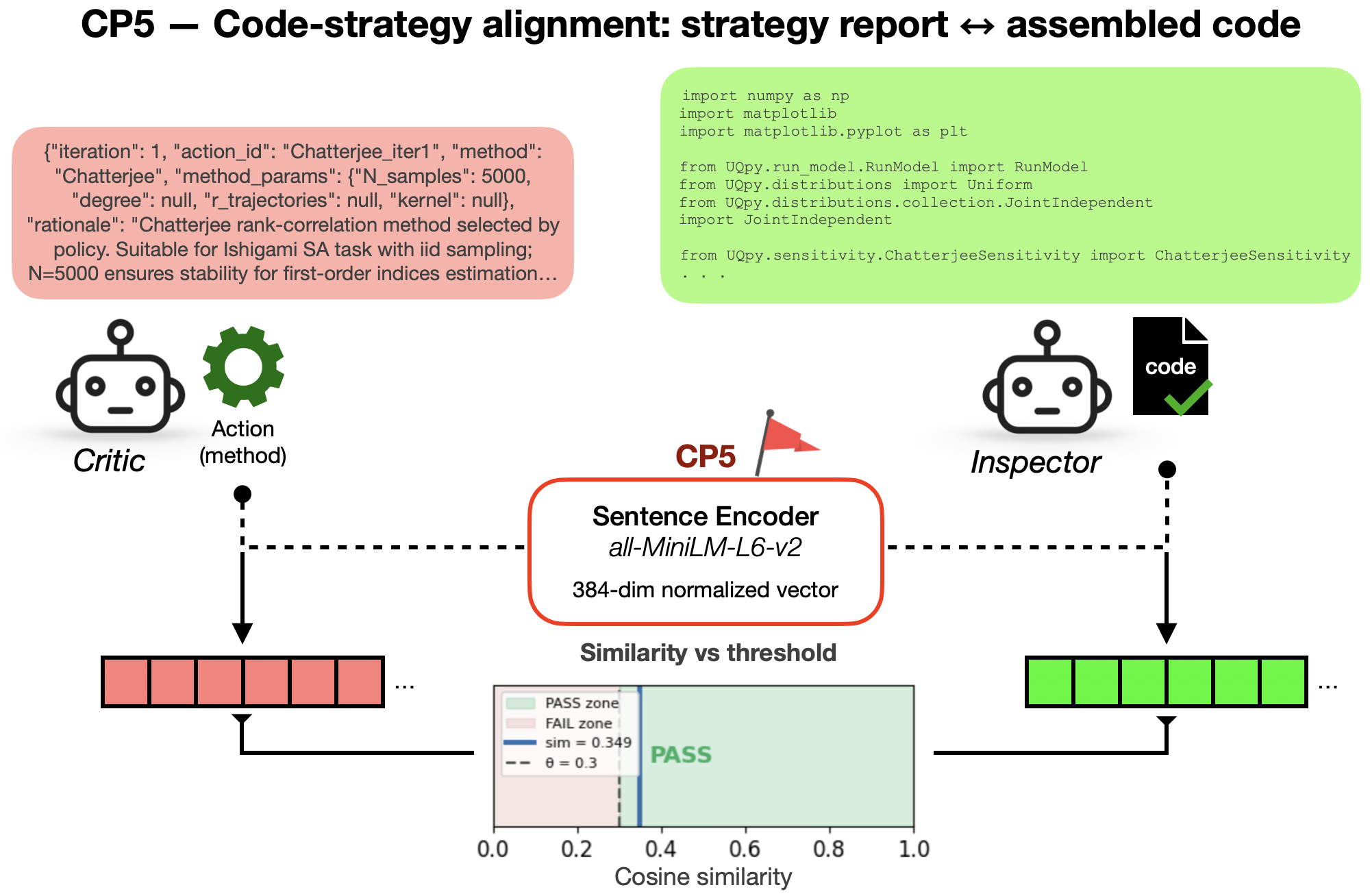}
	   	\caption{Semantic alignment between the approved method (Critic) and the assembled implementation code (Refactor Agent output), verifying that the generated code faithfully reflects the intended method.}
	   	\label{fig:cp5}
   \end{figure}

\newpage
\section{High-Dimensional Screening}\label{sec:exp2}
	
	\begin{figure}[!ht]
		\centering
		\begin{subfigure}[b]{0.95\textwidth}
			\includegraphics[width=1\textwidth]{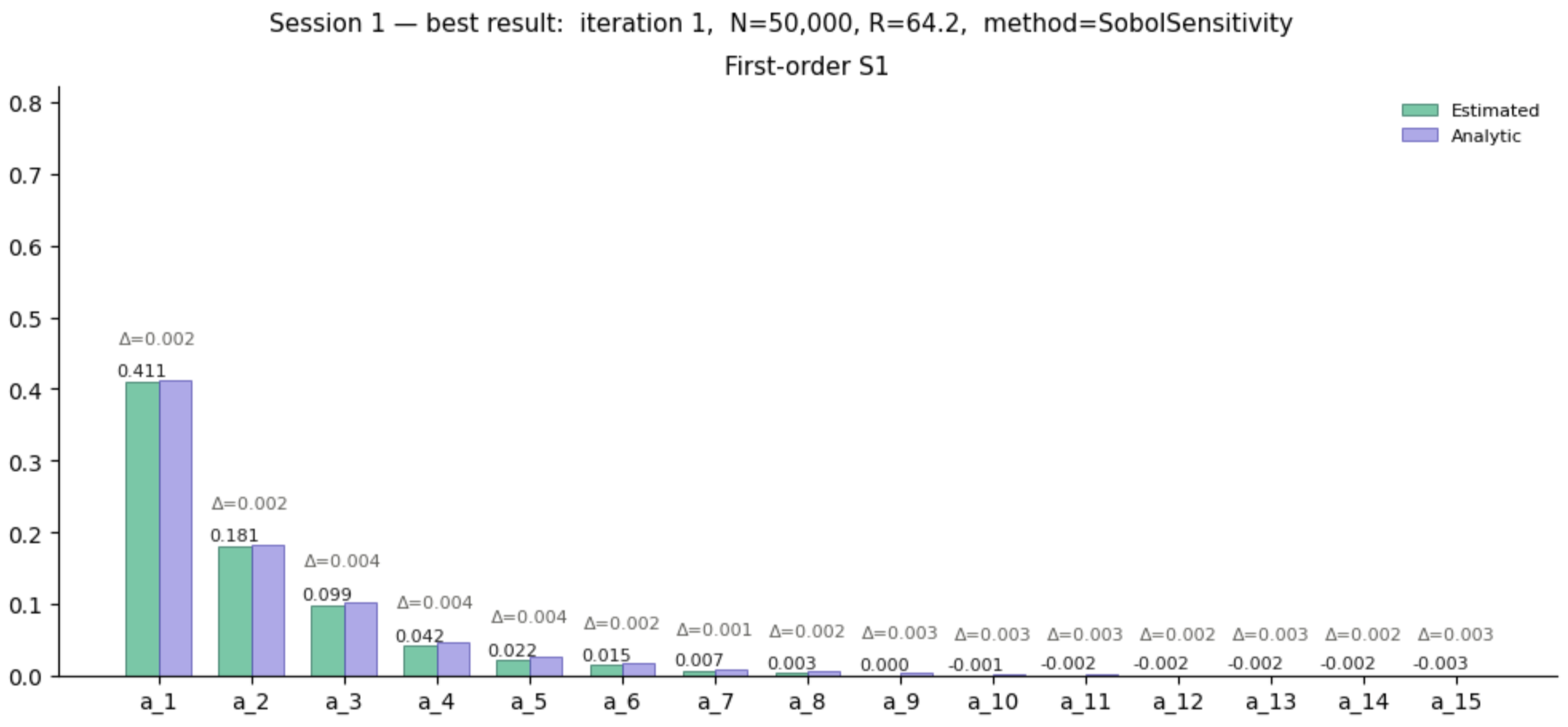}
			\caption{}
		\end{subfigure}
		\begin{subfigure}[b]{0.95\textwidth}
			\includegraphics[width=1\textwidth]{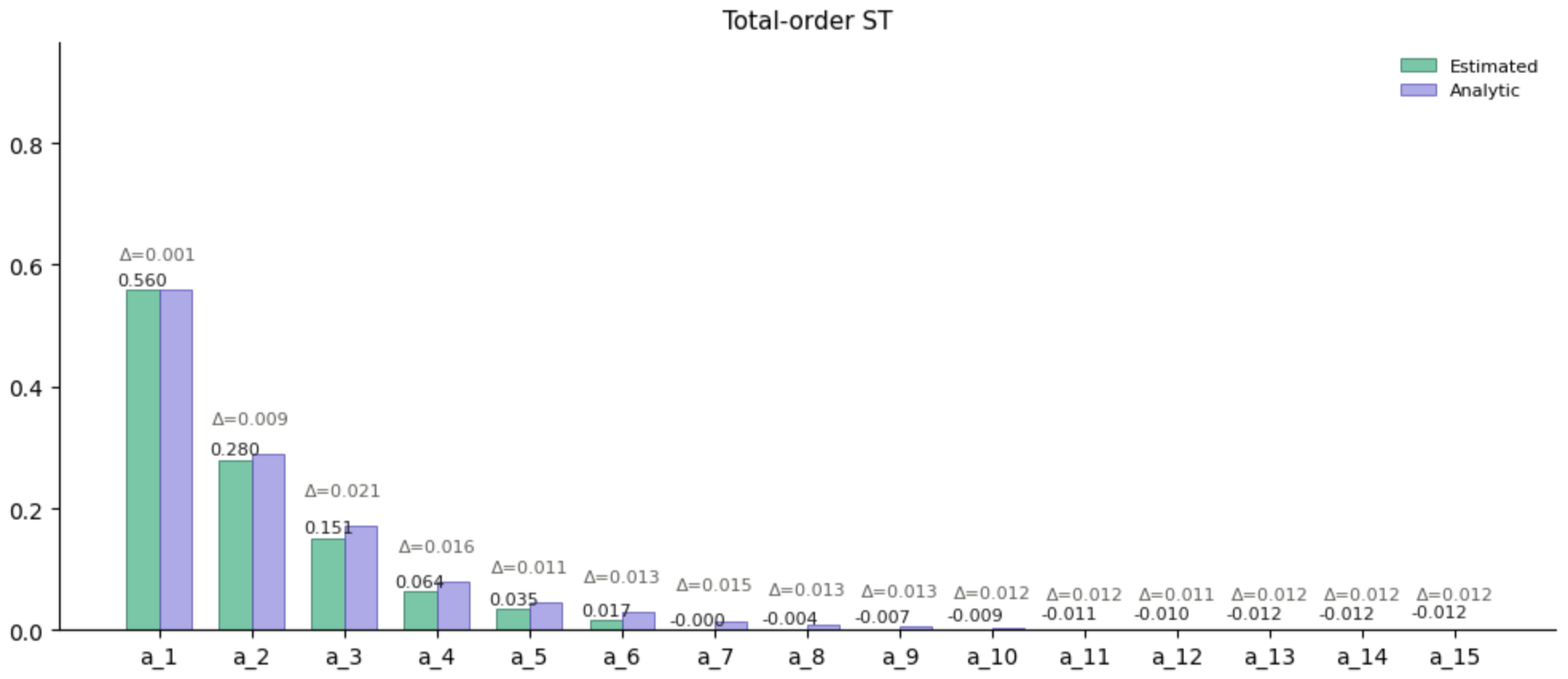}
			\caption{}
		\end{subfigure}
		\begin{subfigure}[b]{0.85\textwidth}
			\includegraphics[width=1\textwidth]{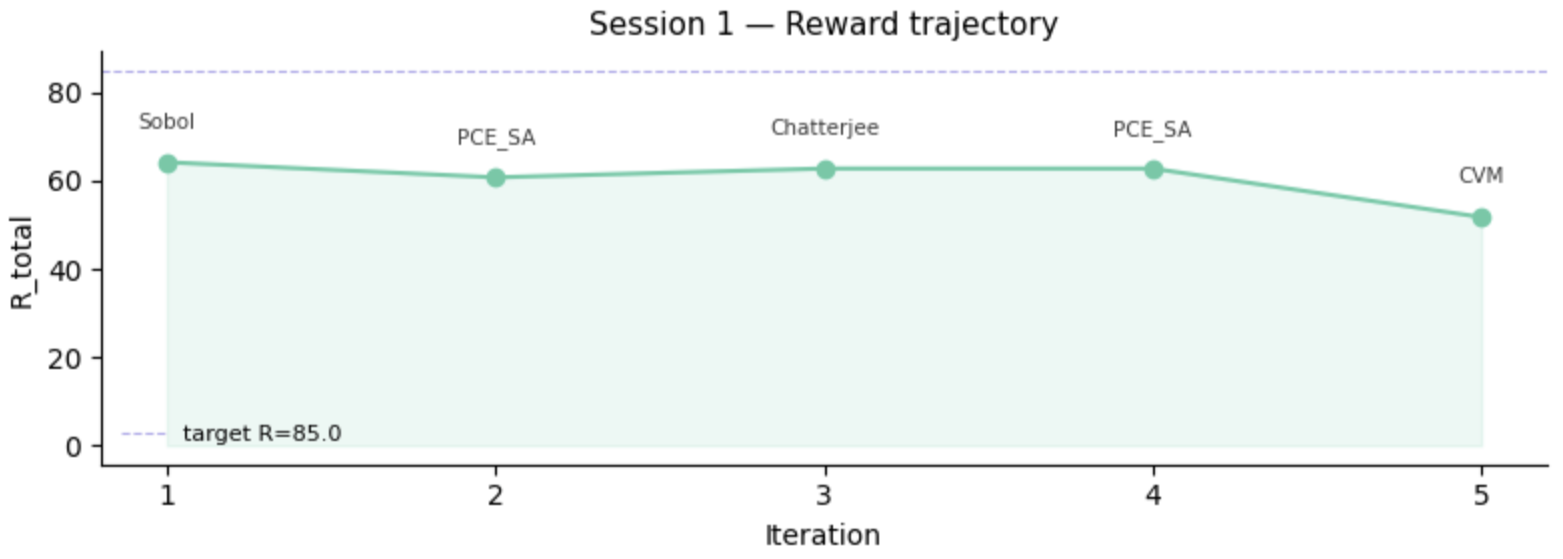}
			\caption{}
		\end{subfigure}
		\caption{}\label{table:GS1}
	\end{figure}
	
	\newpage
	\begin{figure}[!ht]
		\centering
		\begin{subfigure}[b]{0.95\textwidth}
			\includegraphics[width=1\textwidth]{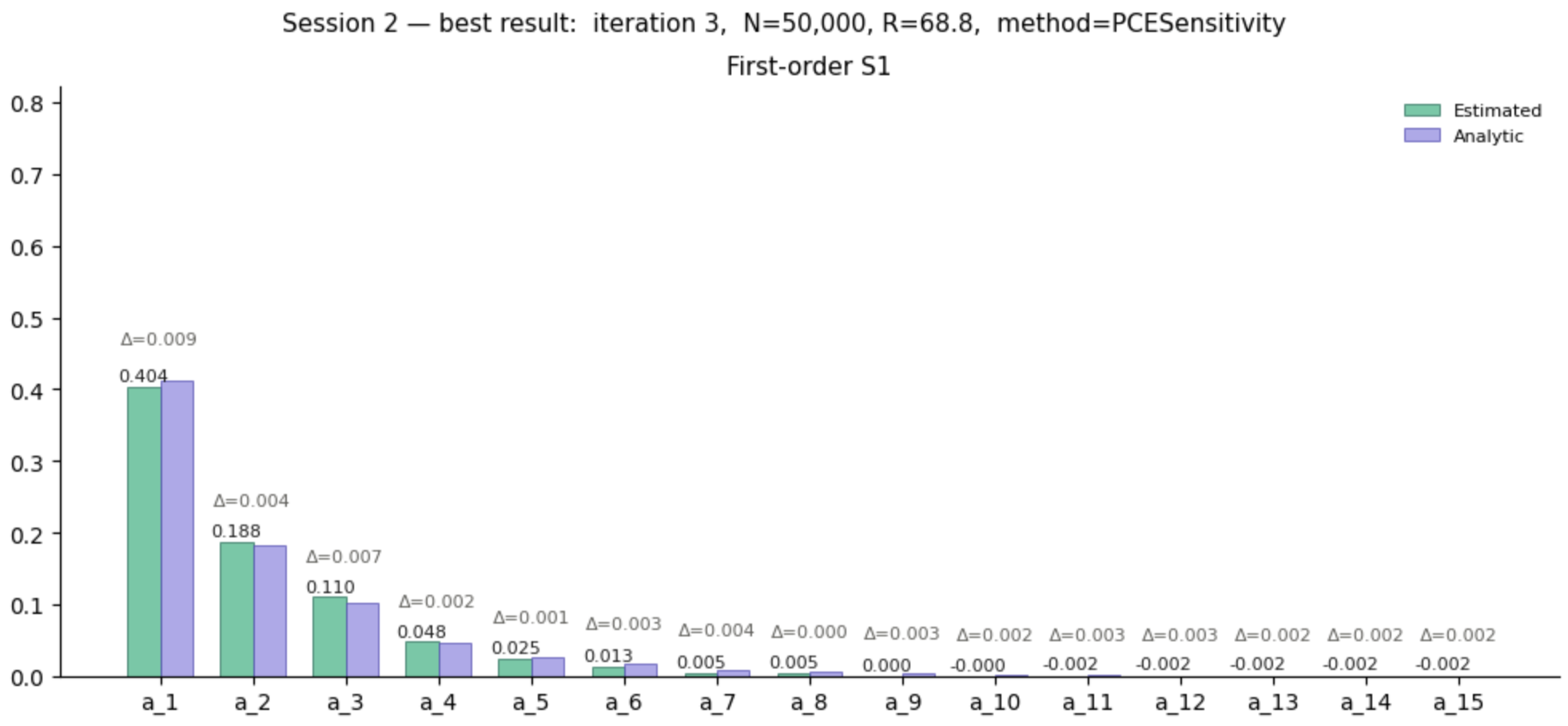}
			\caption{}
		\end{subfigure}
		\begin{subfigure}[b]{0.95\textwidth}
			\includegraphics[width=1\textwidth]{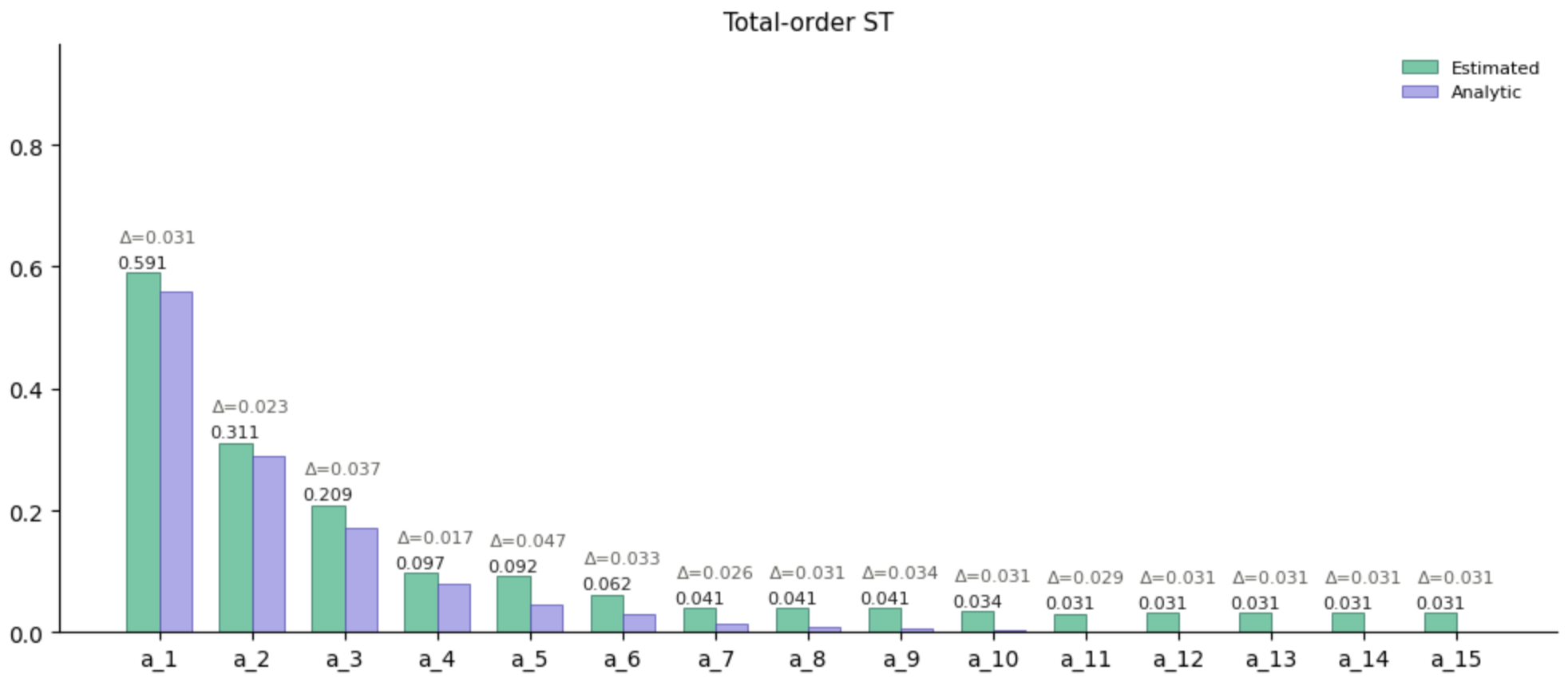}
			\caption{}
		\end{subfigure}
		\begin{subfigure}[b]{0.85\textwidth}
			\includegraphics[width=1\textwidth]{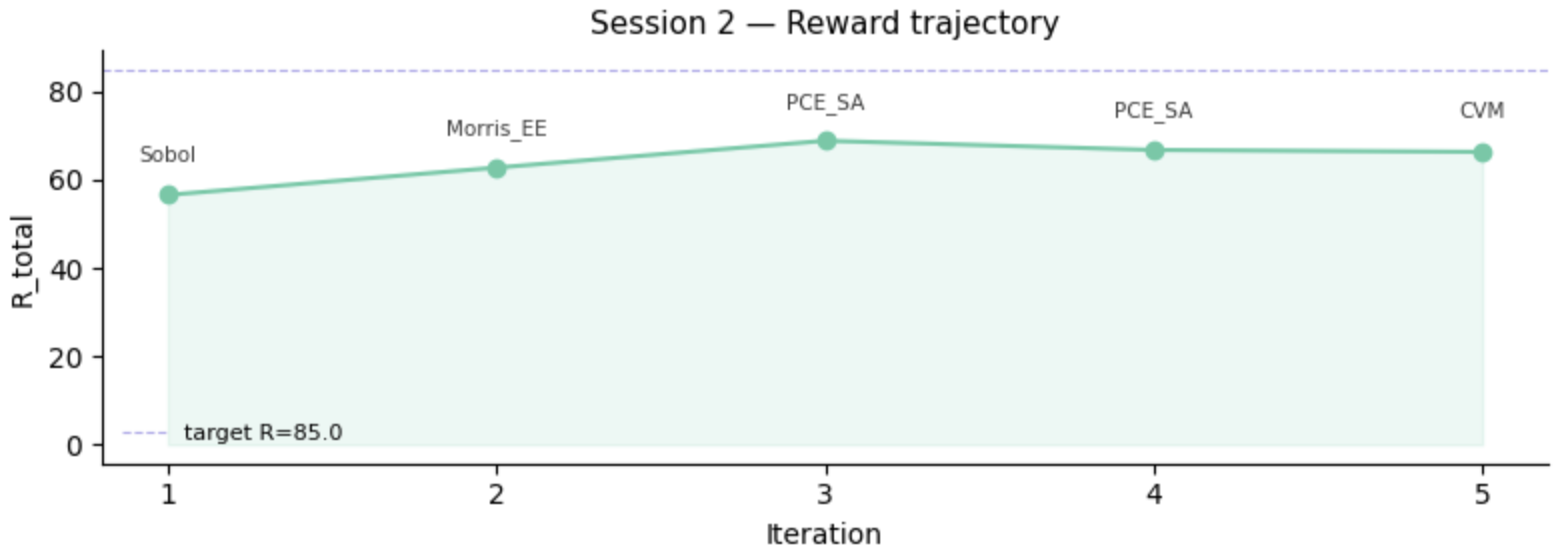}
			\caption{}
		\end{subfigure}
		\caption{}\label{table:GS2}
	\end{figure}
	
	\newpage
	\begin{figure}[!ht]
		\centering
		\begin{subfigure}[b]{0.95\textwidth}
			\includegraphics[width=1\textwidth]{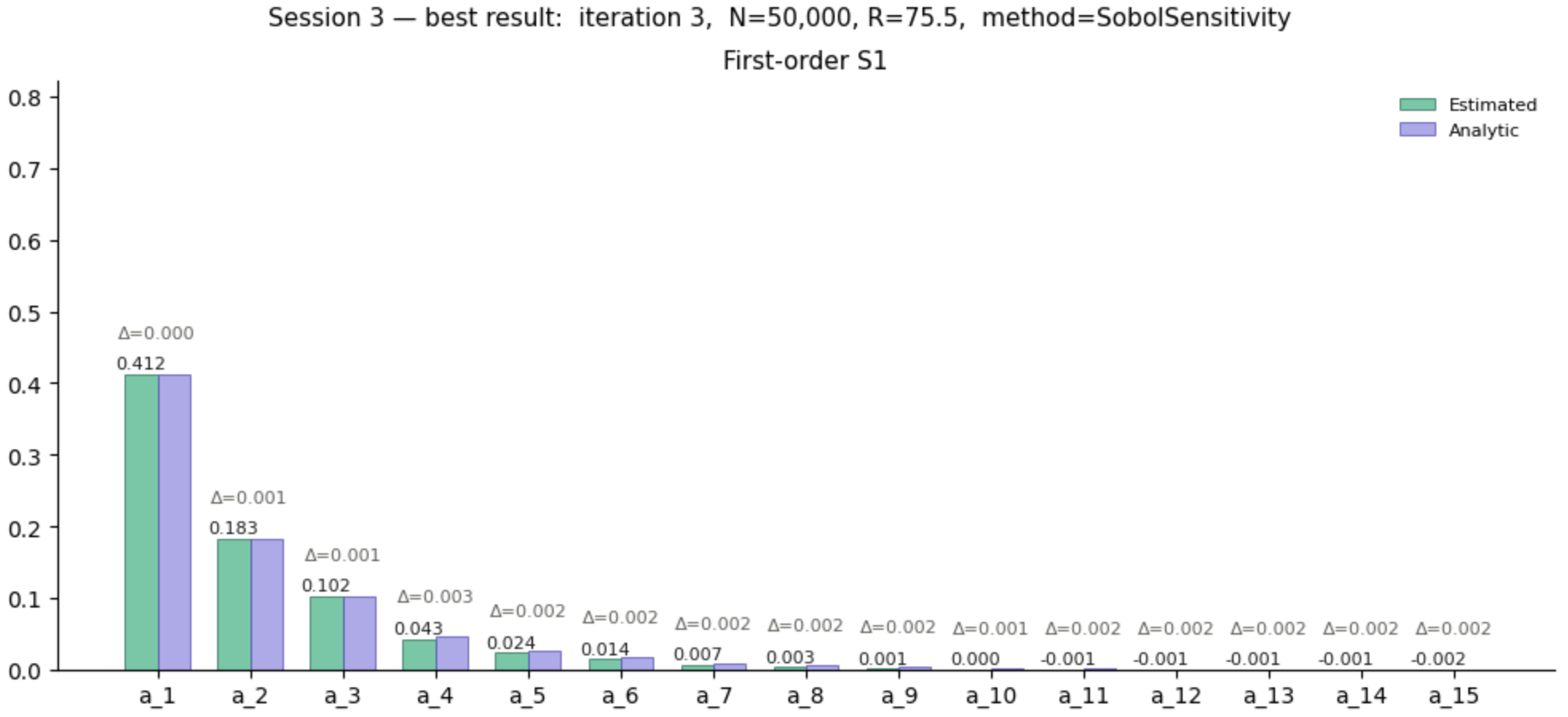}
			\caption{}
		\end{subfigure}
		\begin{subfigure}[b]{0.95\textwidth}
			\includegraphics[width=1\textwidth]{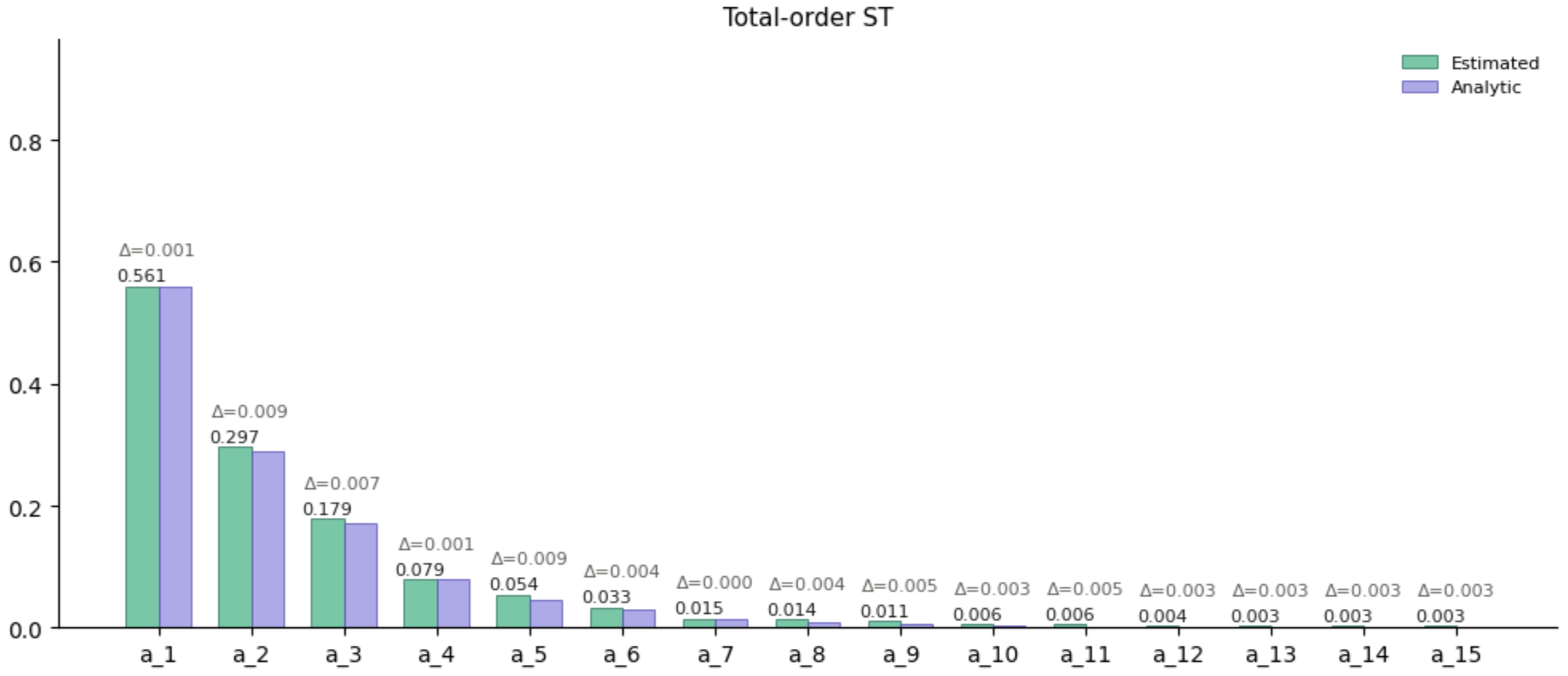}
			\caption{}
		\end{subfigure}
		\begin{subfigure}[b]{0.85\textwidth}
			\includegraphics[width=1\textwidth]{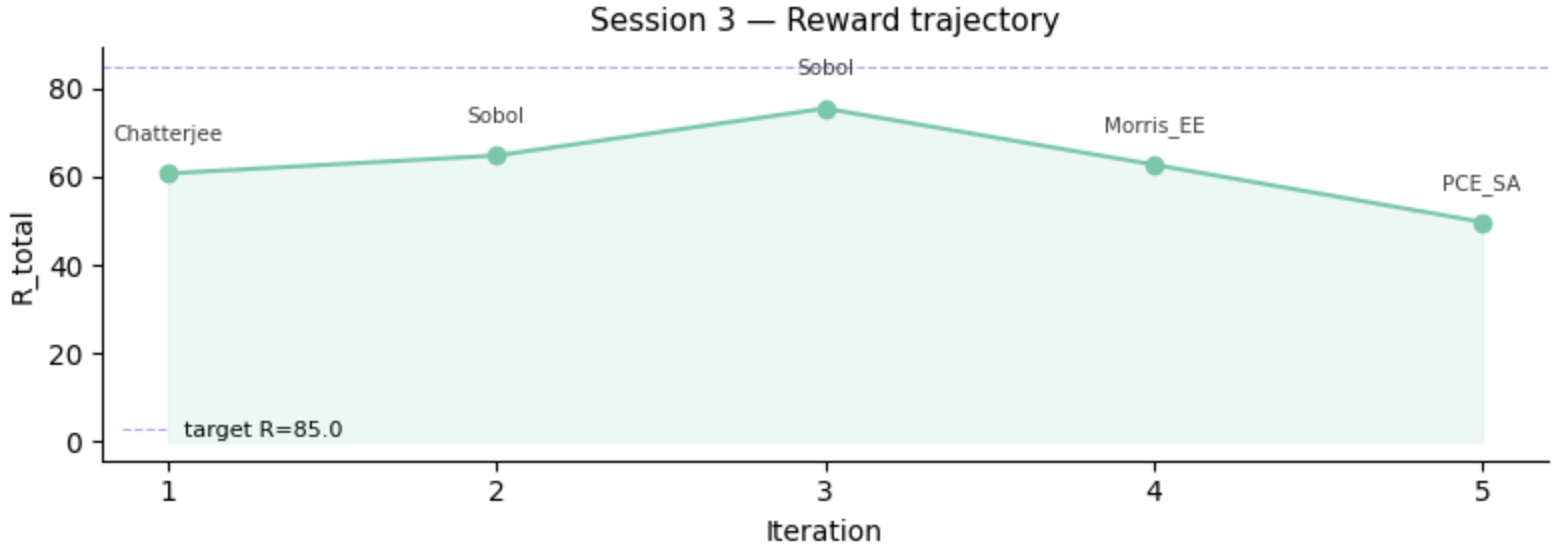}
			\caption{}
		\end{subfigure}
		\caption{}\label{table:GS3}
	\end{figure}

\end{document}